\documentclass{article}
\PassOptionsToPackage{numbers}{natbib}
\usepackage[preprint]{neurips_2026}

\usepackage[utf8]{inputenc} 
\usepackage[T1]{fontenc}    
\usepackage{hyperref}       
\usepackage{url}            
\usepackage{booktabs}       
\usepackage{amsfonts}       
\usepackage{nicefrac}       
\usepackage{adjustbox}
\usepackage{microtype}      
\usepackage[dvipsnames,table]{xcolor}         
\usepackage{graphicx}       
\usepackage{float}          
\usepackage{xspace}         
\usepackage{multirow}
\usepackage{subcaption}
\usepackage{tikz}
\usepackage{forest}
\useforestlibrary{edges}
\usepackage{makecell}
\usepackage{amsmath}
\usepackage{cleveref}      
\usepackage{wrapfig}
\usepackage{array}
\usepackage{anyfontsize}

\usepackage[nolist,nohyperlinks]{acronym}

\begin{acronym}
\acro{llm}[LLM]{Large Language Model}
\acro{mllm}[MLLM]{Multimodal Large Language Model}
\acro{vlm}[VLM]{Vision-Language Model}
\acro{mcq}[MCQ]{Multiple Choice Questions}
\acro{amt}[AMT]{Amazon Mechanical Turk}
\end{acronym}

\newcommand{\cgpt}[1]{GPT-5.4}
\newcommand{\datasub}[1]{label categories}
\newcommand{\qwen}[1]{Qwen3-VL}  
\newcommand{\third}[1]{Gemma~4}  

\newcommand{\siglip}[1]{SigLIP}  
\newcommand{\sigliptwo}[1]{SigLIP2}  
\newcommand{\sigliptwog}[1]{SigLIP2-g}  

\newcommand{\dino}[1]{DINOv3}  
\newcommand{\effnetv}[1]{EffNetV2}  
\newcommand{\effnetl}[1]{EffNet-L2}  
\newcommand{\eva}[1]{EVA-02}  

\newcommand{\gt}[1]{ImGT}
\newcommand{\regt}[1]{ReGT}
\newcommand{\imnet}{ImageNet-1k}

\newcommand{\cntAnoN}{48114}   
\newcommand{\cntReGTS}{31448}
\newcommand{\cntReGTSpos}{28685}
\newcommand{\cntReGTSneg}{2763}
\newcommand{\cntReGTM}{16666}
\newcommand{\cntReGTMpos}{15279}
\newcommand{\cntReGTMneg}{1387}

\newcommand{\attr}[1]{\textit{#1}}

\newcommand{\eg}{\textit{e.g.}\xspace}
\newcommand{\ie}{\textit{i.e.}\xspace}
\newcommand{\cls}[1]{\textup{\texttt{#1}}}

\definecolor{acccolor}{RGB}{52, 168, 83}

\Crefname{section}{Sec.}{Secs.}
\Crefname{figure}{Fig.}{Figs.}
\Crefname{table}{Tab.}{Tabs.}

\title{Doomed to Re-Annotate, Forever: The ImageNet Story}

\author{%
  Illia Volkov \quad Nikita Kisel \quad Tetiana Mishkina \quad
  Klara Janouskova\textsuperscript{*} \quad
  Jiri Matas \\
  Visual Recognition Group, Faculty of Electrical Engineering\\
  Czech Technical University in Prague\\
  \texttt{\{volkoill, kiselnik, mishktet, janoukl1, matas\}@fel.cvut.cz}\\
  \textsuperscript{*}Corresponding author.
}

\begin{document}

\maketitle

\begin{abstract}
Top-1 accuracy on ImageNet-1k remains the most commonly reported metric in visual recognition.
Quality issues with the dataset have been repeatedly 
reported, yet the original 2012 noisy labels are still predominantly used.

The paper presents a comprehensive effort, which goes well beyond prior correction attempts, towards obtaining accurate and complete ImageNet-1k validation set annotations.
The result, \textbf{ReImageNet}, includes multilabel correction, object localization, revised class definitions, and semantic attributes (\attr{text-recognition}, \attr{rendition}, \attr{reflection}, \attr{crowd}, \attr{dominant}).
The reannotation reveals that \textit{$\approx$12\% of the original \imnet{} labels are incorrect}, 
33.3\% of images are multilabel and 3.8\% contain no object from an ImageNet-1k class.
With the new labels, top-1 accuracy increases by  up to $1.2$\% for supervised models
and by  $5$--$6$\%  for MLLMs. 

We argue that annotation at ImageNet scale cannot realistically be completed in one pass, as errors and definitional issues are discovered only through annotating, and we build our pipeline around repeated refinement and error checking.
We observed that human and LLM collaboration with appropriate tooling represents the current quality ceiling for annotation at this scale. 
ImageNet-1k issues propagate into its derivative test sets, indicating that the problem is structural rather than specific to any single benchmark. All annotations, class definitions, guidelines, and analysis code have been publicly released.\footnote{An annotation preview tool is available at \href{https://vrg.fel.cvut.cz/reimagenet}{\nolinkurl{https://vrg.fel.cvut.cz/reimagenet}}.} 
{\small \href{https://huggingface.co/datasets/vrg-prague/ReImageNet}{%
    \raisebox{-0.3em}{\includegraphics[height=1.2em]{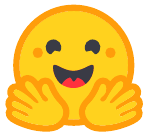}}~Annotations}, \href{https://github.com/klarajanouskova/ImageNet}{%
    \raisebox{-0.3em}{\includegraphics[height=1.2em]{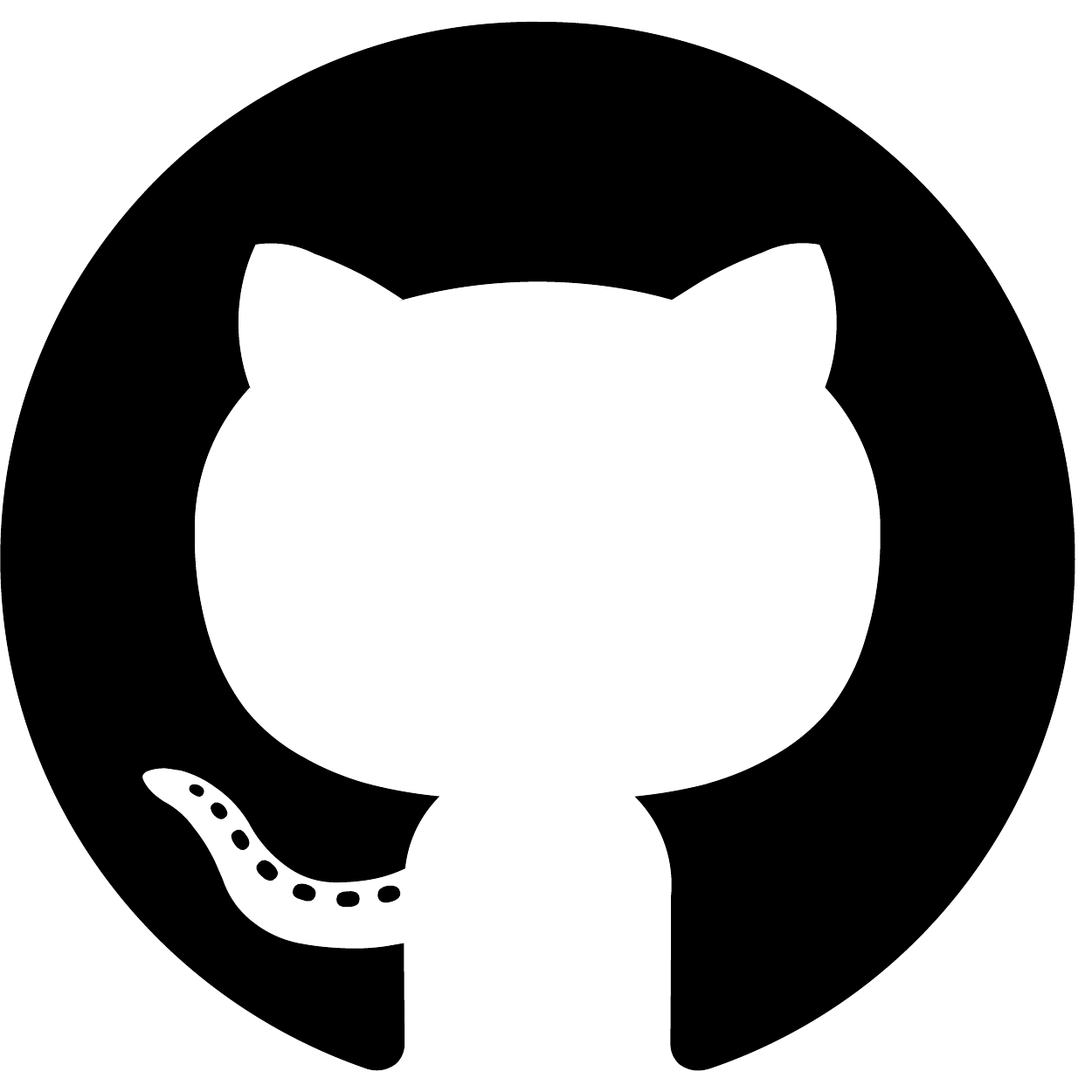}}~Code}.}

\end{abstract}

\section{Introduction}
\label{sec:intro}

Top-1 accuracy on the \imnet{}~\cite{russakovsky2015imagenetlargescalevisual}
validation set remains the single most universally reported metric \cite{rw2019timm} in visual recognition,
even in the era of foundation models and web-scale pretraining.
\imnet{} results appear in virtually every recent vision architecture paper~\cite{hatamizadeh2025mambavisionhybridmambatransformervision, tschannen2025siglip2multilingualvisionlanguage, simeoni2025dinov3, liu2024vmambavisualstatespace, Fang_2024, woo2023convnextv2codesigningscaling, wang2023internimageexploringlargescalevision, liu2022swintransformerv2scaling, tan2021efficientnetv2smallermodelsfaster}
and serve as a core indicator of progress in the field. 

\begin{figure}[ht]
    \centering
    \begin{subfigure}[t]{0.4\linewidth}
        \centering
        \includegraphics[width=0.96\linewidth]{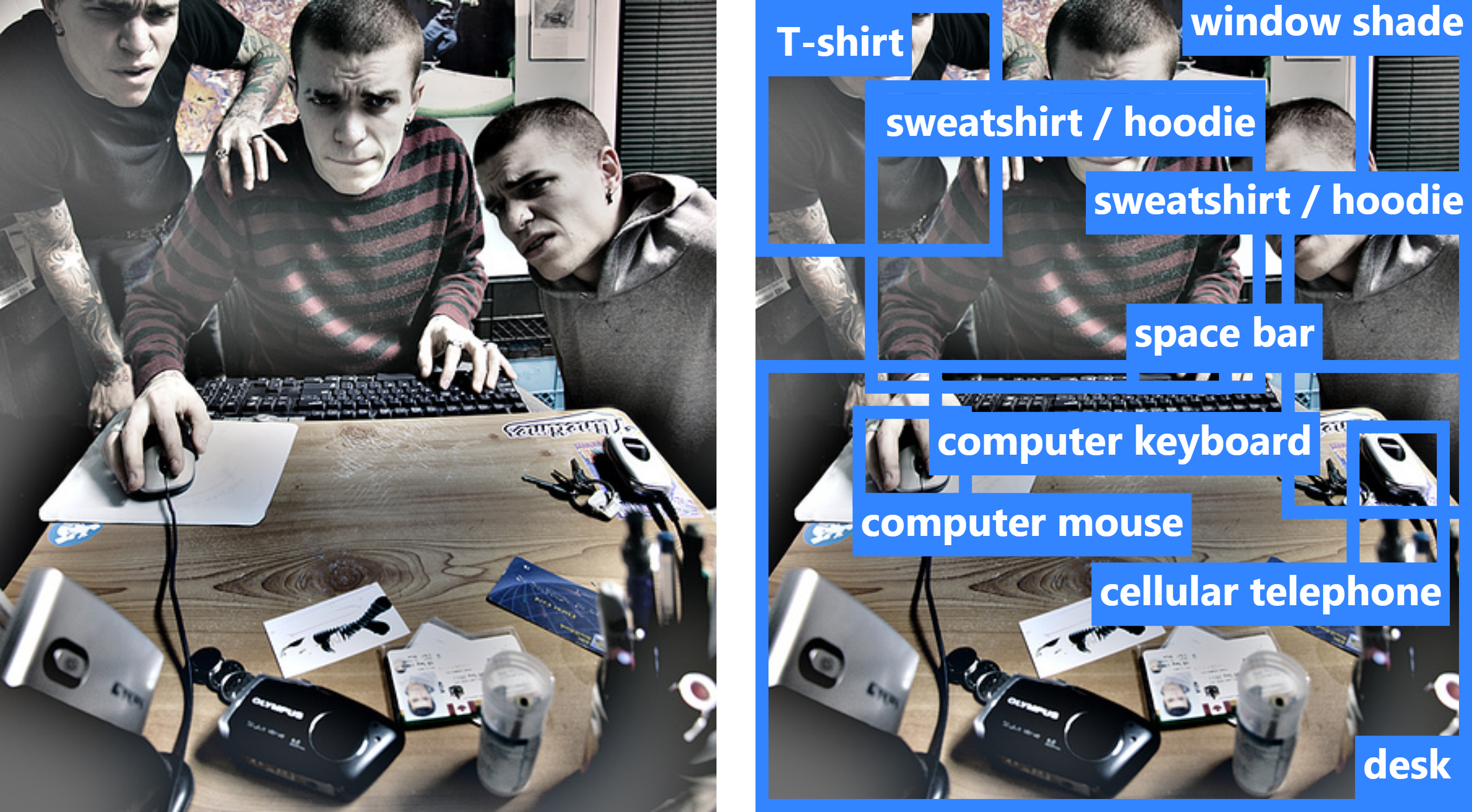}
        \caption{\raggedright
        \gt{}: \textcolor{ForestGreen}{computer mouse} \\
        \hphantom{(a) }\regt{}: \textcolor{ForestGreen}{computer mouse}; \textcolor{ForestGreen}{space bar}; \textcolor{ForestGreen}{desk};\\
        \hphantom{(a) }\textcolor{ForestGreen}{sweatshirt~/~hoodie}; \textcolor{ForestGreen}{computer keyboard}; \\
        \hphantom{(a) }\textcolor{ForestGreen}{cellular telephone};  \textcolor{ForestGreen}{T-shirt};   \textcolor{ForestGreen}{window shade}
        }
    \end{subfigure}
    \hfill
    \begin{subfigure}[t]{0.58\linewidth}
        \centering
        \includegraphics[width=\linewidth]{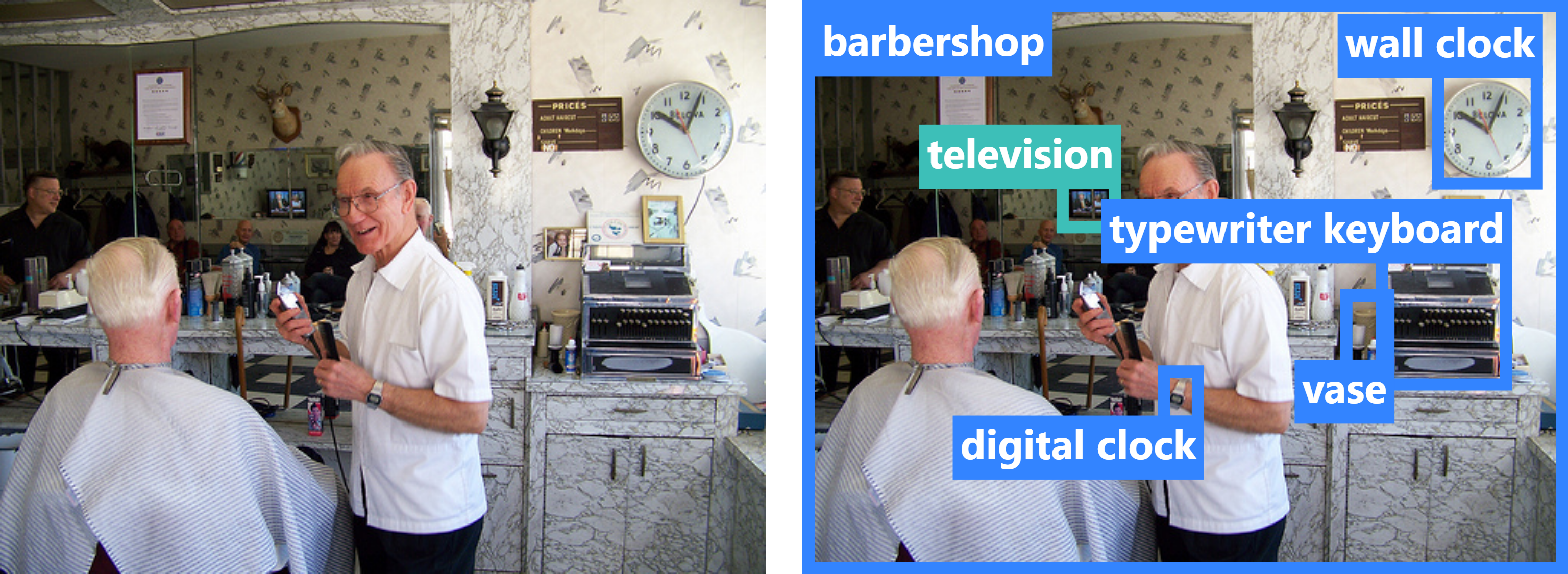}
        \caption{\raggedright
        \gt{}: \textcolor{ForestGreen}{barbershop} \\
        \hphantom{(a) }\regt{}: \textcolor{ForestGreen}{barbershop}; \textcolor{ForestGreen}{television}; \textcolor{ForestGreen}{typewriter keyboard}; \textcolor{ForestGreen}{vase} \\ 
        \hphantom{(a) }\textcolor{ForestGreen}{wall clock}; \textcolor{ForestGreen}{digital clock}; 
        }
    \end{subfigure}

    \vspace{0.5em}

    \begin{subfigure}[t]{0.35\linewidth}
        \centering
        \includegraphics[width=0.93\linewidth]{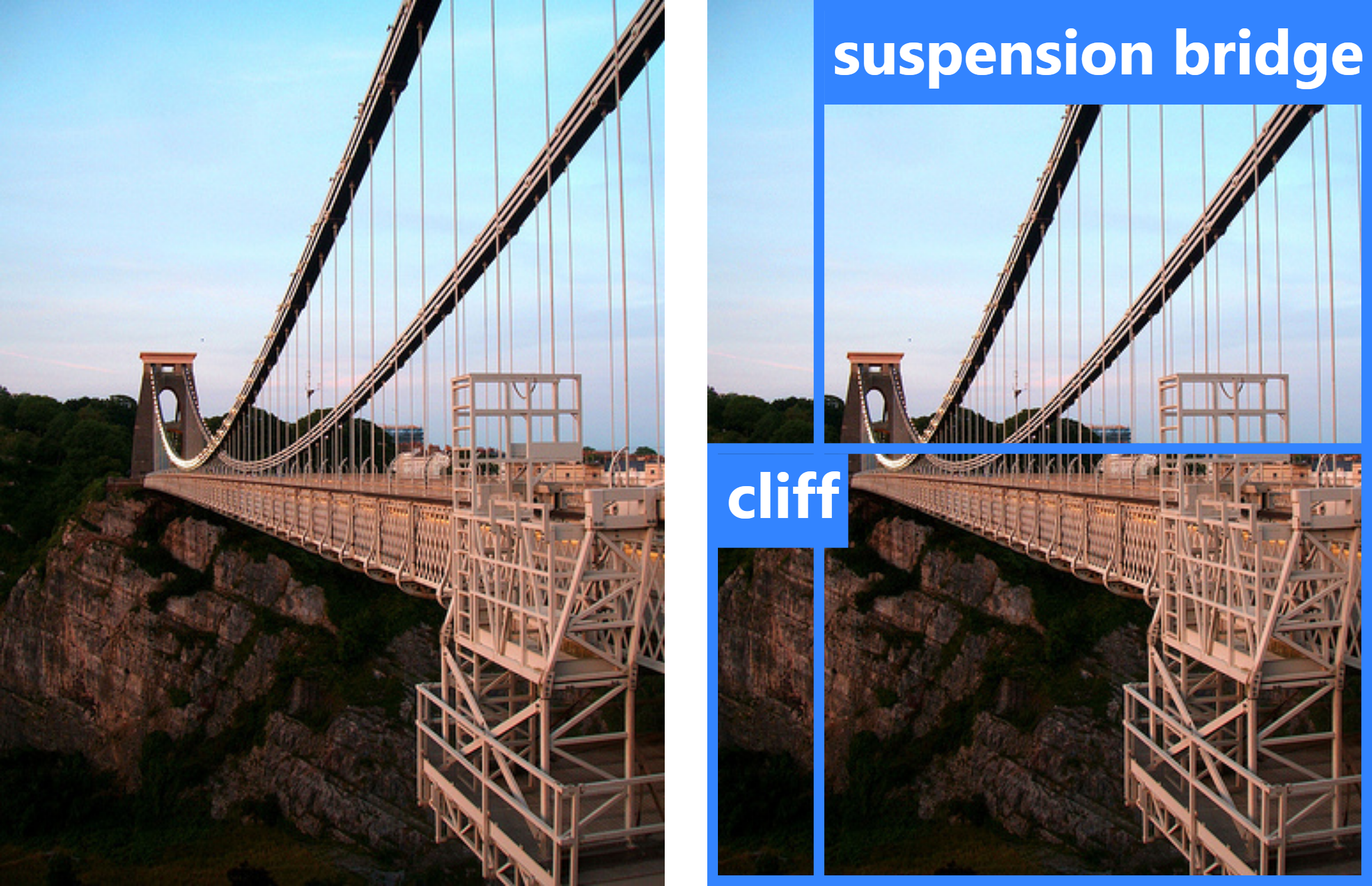}
        \caption{\raggedright
        \gt{}: \textcolor{red}{pier} \\
        \hphantom{(a) }\regt{}: \textcolor{ForestGreen}{suspension bridge}; \textcolor{ForestGreen}{cliff}
        }
    \end{subfigure}
    \hfill
    \begin{subfigure}[t]{0.64\linewidth}
        \centering
        \includegraphics[width=\linewidth]{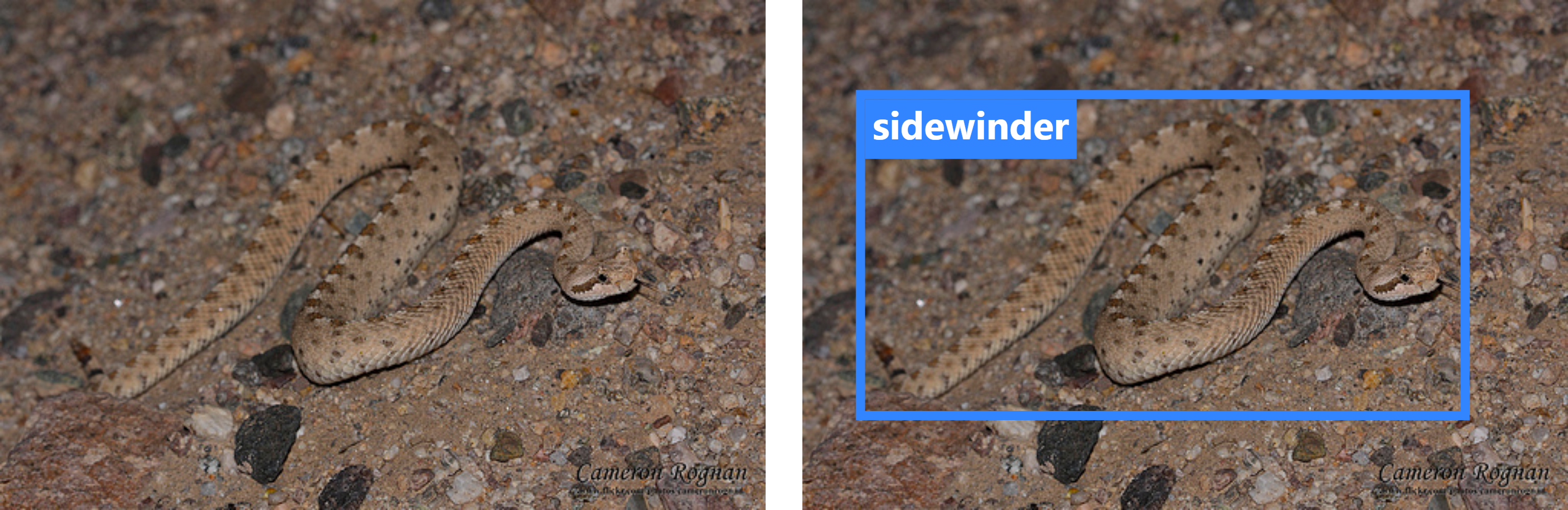}
        \caption{\raggedright
        \gt{}: \textcolor{red}{Saharan horned viper} \\
        \hphantom{(a) }\regt{}: \textcolor{ForestGreen}{sidewinder}
        }
    \end{subfigure}

    \caption{
    \gt{} -- original single-label annotations
    (left) and \regt{} -- new multilabel localized reannotations (right). 
    For both (a) and (b), \gt{} is correct but incomplete -- \regt{} labels at least all objects a person notices within 5 seconds. In~(b), \cls{television} is a \textcolor[HTML]{3DBFB8}{\rule{1.2ex}{1.2ex}}~\attr{reflection} visible in the mirror (see~\Cref{fig:attributes}). Image (c)~shows an obvious labelling error.
    Image (d)~needed a fine-grained correction: the two snake species are visually near-identical to non-experts due to convergent evolution; the \regt{} was confirmed via reverse image search, tracing the photograph to a Utah-based wildlife biologist.}
    \label{fig:teaser}
\end{figure}


Yet \imnet{} is far from clean: a growing body of 
work has documented label errors~\cite{northcutt2021pervasivelabelerrorstest}, 
omissions of multiple labels~\cite{beyer2020imagenet, pmlr-v119-shankar20c, 
tsipras2020imagenetimageclassificationcontextualizing}, 
class overlaps~\cite{vasudevan2022doesdoughbagelanalyzing}, 
and deeper structural problems, including a distribution shift between 
training and validation sets~\cite{kisel2024flaws}.
These issues have consequences: label noise introduces ranking 
instability affecting which models are selected for downstream 
use. 
Prior validation set reannotation efforts~\cite{beyer2020imagenet, pmlr-v119-shankar20c, 
tsipras2020imagenetimageclassificationcontextualizing} made
progress, but followed the original, problematic class taxonomy~\cite{kisel2024flaws}.
As a result, the corrected labels from different papers are 
inconsistent~\cite{kisel2024flaws}.
These efforts also came before the availability of powerful \acp{mllm}, 
which make exhaustive multilabel annotation more manageable at this 
scale~\cite{kisel2026multimodallargelanguagemodels, xie2026multimodallargelanguagemodels}.


In the paper, we describe a large-scale effort to obtain accurate, complete, and, wherever possible, backward-compatible ImageNet-1k validation set annotations. 
Towards this end, we go well beyond the prior art. 
We train in-house annotators and provide them classes in semantically coherent groups derived from the WordNet \cite{miller-1994-wordnet} hierarchy, reducing the cognitive burden of discriminating among 1{,}000  classes simultaneously. 
Work on each group starts with a familiarisation pass: annotators study the images and 
revise class definitions and names where needed.
Then they proceed to per-image labeling,  which is supported by a detector, \ac{vlm}, and \ac{mllm} predictions. 
Annotators label the objects that stand out on first viewing, not every object they can find. 
Our initial protocol asked for the latter, but objects spanning only a few dozen pixels, recognisable only from context and after zooming in, made the label set depend on how hard each annotator chose to look.

\begin{figure}[t]
    \centering
    \fontsize{7.4}{9.5}\selectfont
    \setlength{\tabcolsep}{4pt}
    \begin{tabular}{p{0.21\linewidth} p{0.22\linewidth} p{0.21\linewidth} p{0.28\linewidth}}
        
        \parbox[c]{\linewidth}{\includegraphics[width=\linewidth]{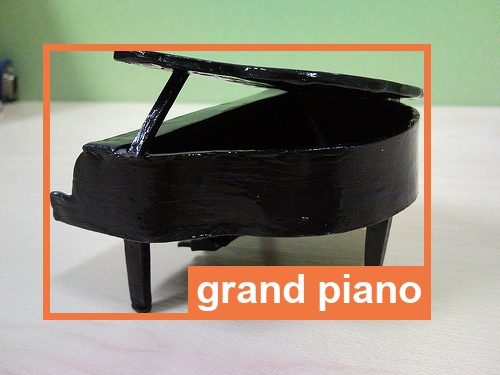}} &
        \parbox[c]{\linewidth}{\textcolor[HTML]{F07840}{\rule{1.2ex}{1.2ex}}~\attr{Rendition} --- the depicted object visually mimics a class instance but appears as a toy, drawing, or other artificial or stylized representation. Useful for measuring model generalisation and robustness.} &
        \parbox[c]{\linewidth}{\includegraphics[width=\linewidth]{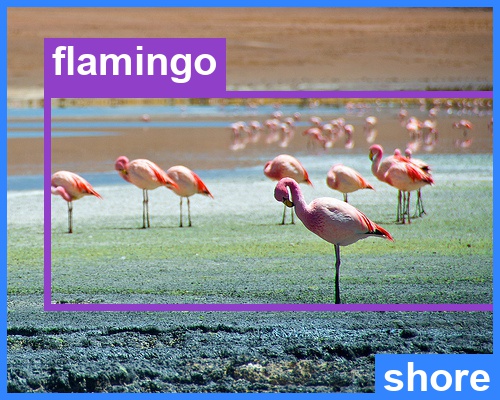}} &
        \parbox[c]{\linewidth}{\textcolor[HTML]{9040C8}{\rule{1.2ex}{1.2ex}}~\attr{Crowd} --- five or more instances of the same class appear in the image. To reduce annotation effort, they are collapsed into a single bounding box. The exact instance count is not recorded, so instance-level counting is not supported for images with this attribute.} \\[4.5em] 
        
        \parbox[c]{\linewidth}{\includegraphics[width=\linewidth]{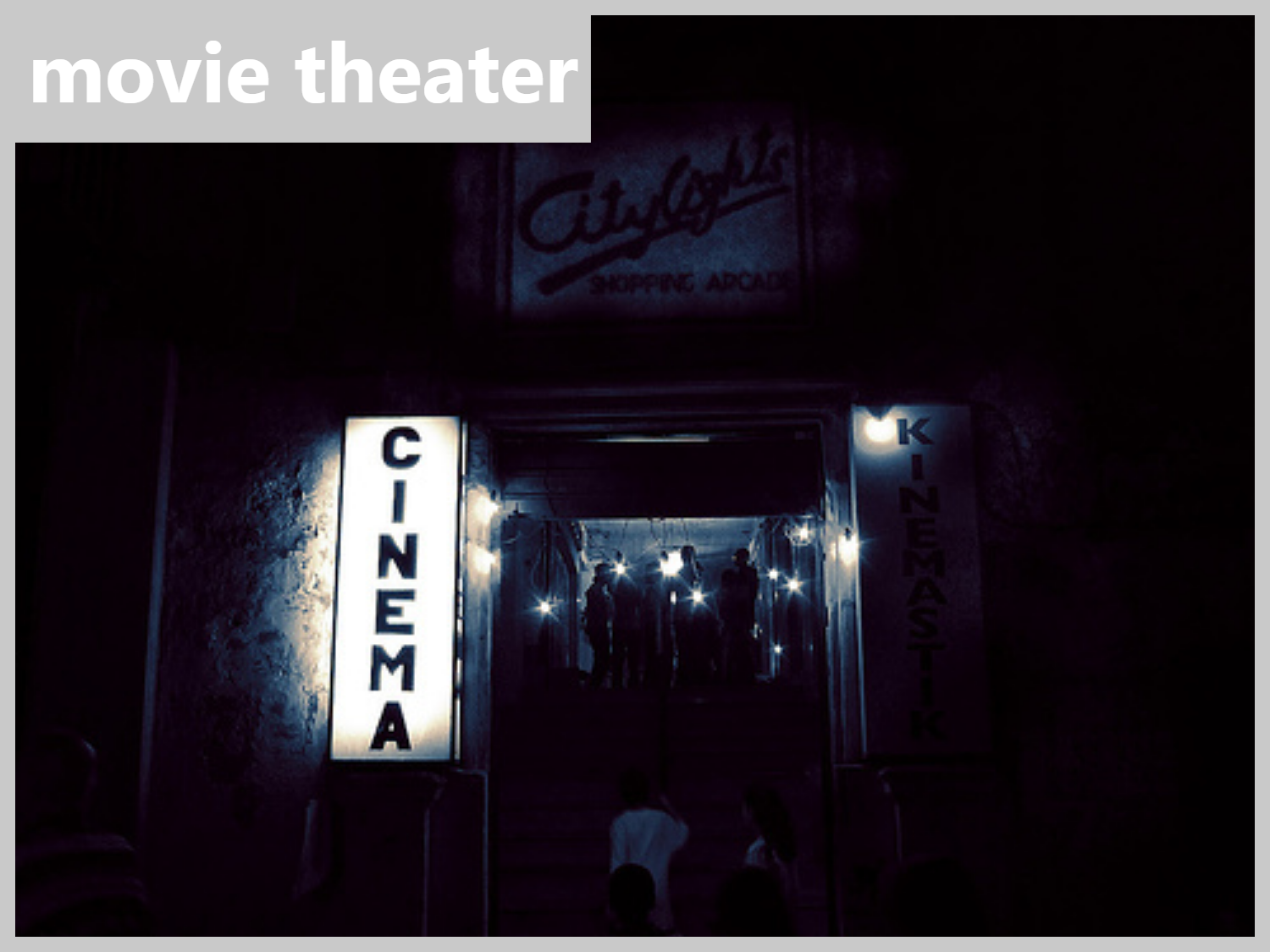}} &
        \parbox[c]{\linewidth}{\textcolor[HTML]{C9C9C9}{\rule{1.2ex}{1.2ex}}~\attr{Text-recognition} --- classification of the object requires reading text visible in the image. Tests whether models can leverage visible text as a recognition cue.} &
        \parbox[c]{\linewidth}{\includegraphics[width=\linewidth]{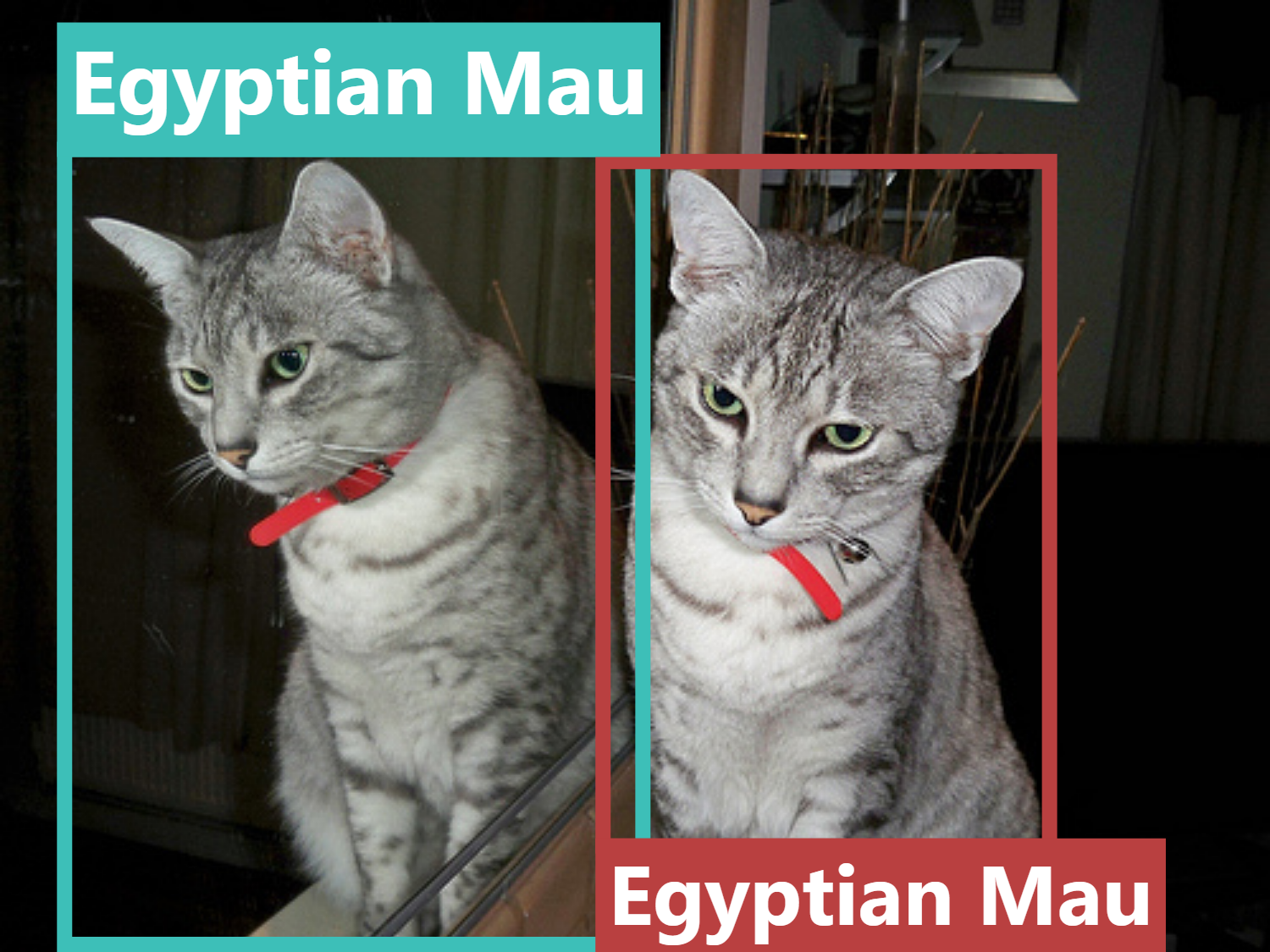}} &
        \parbox[c]{\linewidth}{\textcolor[HTML]{3DBFB8}{\rule{1.2ex}{1.2ex}}~\attr{Reflection} --- 
        in a mirror or water surface, annotated independently of whether the reflected object itself is also visible in the image. Measures model robustness to reflected appearances and affects instance counting.} \\[4.5em]
        
        \multicolumn{4}{l}{
            \parbox[c]{0.16\linewidth}{\includegraphics[width=\linewidth]{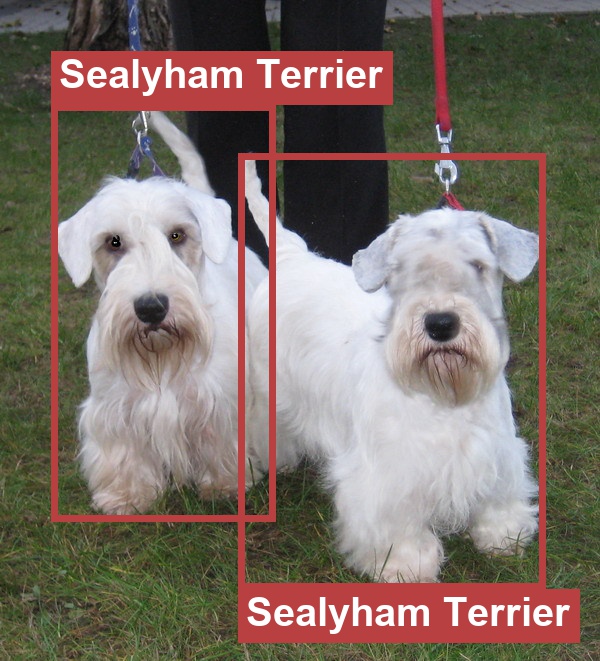}}
            \hspace{0.3em}
            \parbox[c]{0.23\linewidth}{\includegraphics[width=\linewidth]{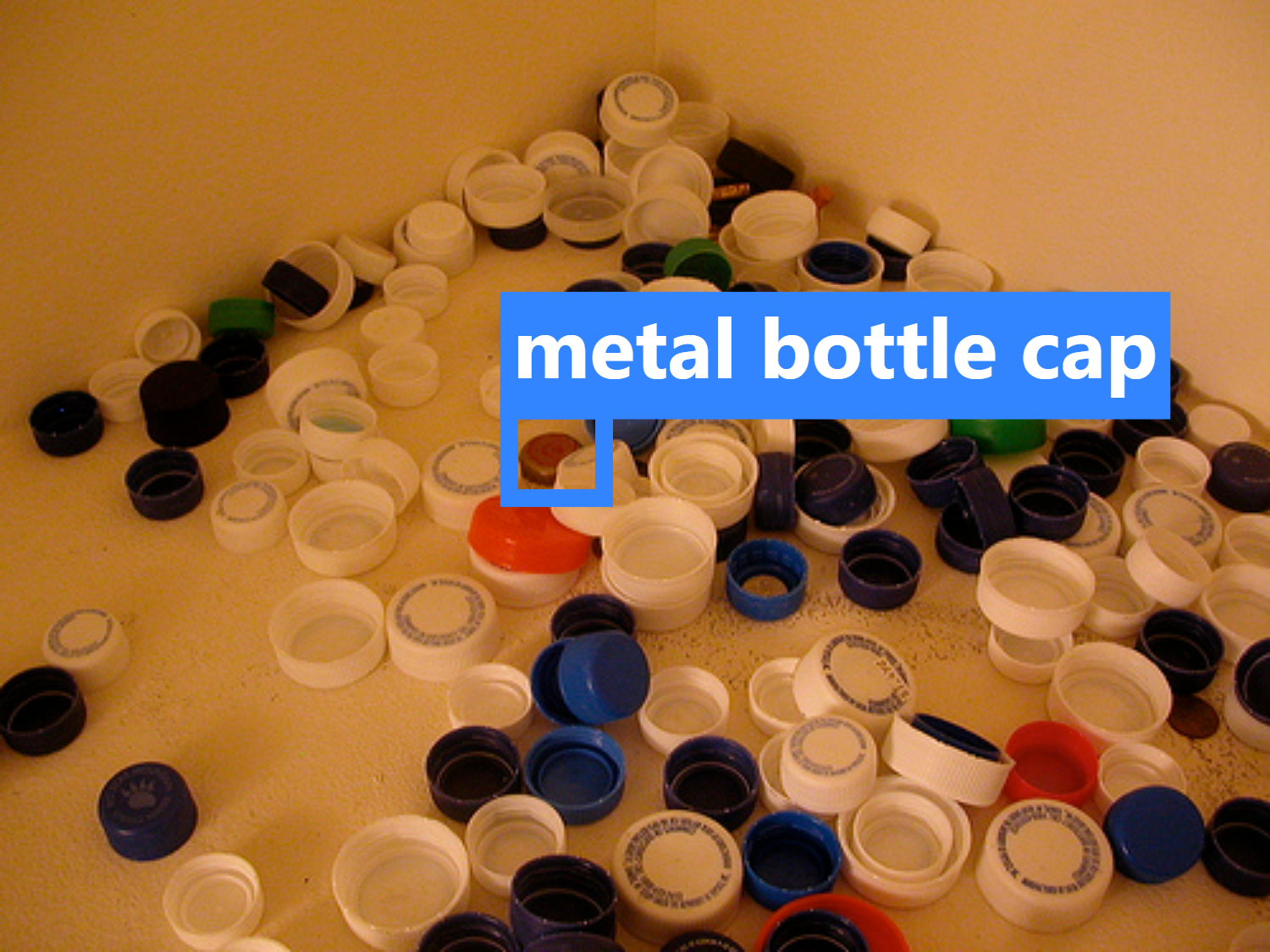}}
            \hspace{0.5em}
            \parbox[c]{0.55\linewidth}{\textcolor[HTML]{B94040}{\rule{1.2ex}{1.2ex}}~\attr{Dominant} --- assigned to objects that a person notices immediately upon viewing the image, based on annotator judgment rather than size alone. An image may contain any number of dominant objects (\eg both \cls{Sealyham Terrier} instances on the left are \attr{dominant}). An object may also be the only ImageNet class in the image and still not be dominant (\eg~\cls{metal bottle cap} on the right).\\[0.5em]
            \textcolor[HTML]{3384FF}{\rule{1.2ex}{1.2ex}}~Denotes objects with no \attr{attribute}.}
        } \\
    \end{tabular}
    \caption{Annotation extensions: \attr{attributes}.
    Objects may have multiple, only one is visualized here.}
    \label{fig:attributes}
    
\end{figure}

Continuous quality checks and active communication with and among annotators were uncovering edge cases throughout the process, requiring repeated mid-annotation revisions of the guidelines and class definitions. This resulted in the introduction of {attributes}: \attr{rendition}, 
\attr{reflection}, and \attr{text-recognition} to consistently handle recurring 
questions like "is a toy gun a toy (or a gun)?" and "is an image of a face in a mirror a face?" (see~\Cref{fig:attributes}). Attributes \attr{crowd} and \attr{dominant} were introduced to reduce 
the load in images with many objects of the same or 
different classes, respectively.

Many images in \imnet{} depict multiple instances from several classes.
For each image, we aim to provide bounding boxes for at least each immediately visible instance.
Multilabel evaluation, however, is permissive: an image with several valid 
labels is counted correct for any of them, so accuracy on \regt{} is not 
directly comparable to the single-label numbers reported over the past decade.
We therefore also evaluate on expanded object crops, each containing exactly 
one annotated object, which recovers a strict single-label protocol on 
localised annotations (see~\Cref{subsec:crops}).

{\bf Limitations.} Several sources of bias are inherent to our pipeline: annotators 
share a European background and are non-experts in fine-grained 
domains, model predictions may nudge labelling decisions, the effort 
covers only the validation set, and the images themselves date from 
2012 and may no longer represent current instances of some classes. These are discussed in detail 
in~\Cref{subsec:limitations}.

The challenges we document are not unique to \imnet{} but represent typical failure modes of large-scale visual benchmarks constructed under time and budget pressure. In~\Cref{subsec:derived}, we show that annotation errors in \imnet{} propagate directly into its derivative test sets  \cite{hendrycks2021naturaladversarialexamples,hendrycks2021many,wang2019learningrobustglobalrepresentations,recht2019imagenetclassifiersgeneralizeimagenet,NEURIPS2019_97af07a1,wang2024soberlookrobustnessclips}. 

\noindent\textbf{The contributions of the paper are:} 
\textbf{(i)} ReImageNet, a reannotated \imnet{} validation set with multilabel corrections, localized bounding boxes, revised class definitions, and rich semantic attributes enabling counting and detection evaluation; 
\textbf{(ii)} an  evaluation of model performance on the corrected labels, 
label count analysis, and crop-based object-centric evaluation protocols;
\textbf{(iii)} analysis of annotation error propagation to derivative \imnet{} benchmarks;
and \textbf{(iv)} a characterization of the challenges facing large-scale visual benchmark construction that extends beyond \imnet{}, with practical guidance on the iterative high-quality annotation and human-\ac{mllm} collaboration.

\section{Related Work}
\label{sec:background}

\textbf{\noindent{Label Quality of ImageNet.}} ImageNet-1k has been a key benchmark in computer vision, but continued community review has uncovered problems that reduce its reliability. 
Several reannotation efforts have improved the label quality, mainly in the validation set: \cite{pmlr-v119-shankar20c} carried 
out focused relabeling with multi-label handling; 
\cite{tsipras2020imagenetimageclassificationcontextualizing} used multiple 
annotators per image on a selected subset; \cite{beyer2020imagenet} 
expanded this to the full validation set and introduced the ReaL accuracy 
metric; \cite{northcutt2021pervasivelabelerrorstest} estimated label error 
rates across popular benchmarks. Automatic relabeling of the training set has also been explored  \cite{beyer2020imagenet, 
yun2021relabelingimagenetsinglemultilabels}. 

Later work shifted to 
deeper structural problems: \cite{vasudevan2022doesdoughbagelanalyzing} 
studied class overlap and duplicate images; \cite{kisel2024flaws} identified a distribution shift between training and validation sets, class name mismatches with actual image content, near-synonym classes, and the subclass--superclass overlaps. Building on this, we 
carried out a partial reannotation that produced measurable changes in 
model rankings \cite{kisel2026multimodallargelanguagemodels}. In this work, we annotate the 
full validation set from scratch, unifying multilabel correction, localisation, 
revised class definitions, and structured attribute annotation in a single 
effort.

\textbf{\noindent{Derivative Benchmarks and Robustness Evaluation.}} 
Many benchmarks build on \imnet{} \cite{moayeri2022hard, hendrycks2019robustness, liu2019largescalelongtailedrecognitionopen, recht2019imagenetclassifiersgeneralizeimagenet, hendrycks2021naturaladversarialexamples, hendrycks2021many, wang2019learningrobustglobalrepresentations, NEURIPS2019_97af07a1, wang2024soberlookrobustnessclips, idrissi2022imagenetxunderstandingmodelmistakes}, inheriting either its images and labels 
or its taxonomy, resulting in its problems propagating to 
them~(see~\Cref{subsec:derived}). The first group reuses the original images 
and directly inherits the annotation noise~(\eg Hard 
ImageNet~\cite{moayeri2022hard}, ImageNet-C/-P~\cite{hendrycks2019robustness}, 
ImageNet-LT~\cite{liu2019largescalelongtailedrecognitionopen}). The second 
adopts the \imnet{} taxonomy on newly collected images. 
ImageNetV2~\cite{recht2019imagenetclassifiersgeneralizeimagenet} closely 
mirrors the original collection process, so structural issues carry over at a 
similar rate. Others target specific robustness 
aspects~(ImageNet-A~\cite{hendrycks2021naturaladversarialexamples}, 
ImageNet-R~\cite{hendrycks2021many}, 
ImageNet-Sketch~\cite{wang2019learningrobustglobalrepresentations}, 
ObjectNet~\cite{NEURIPS2019_97af07a1}, 
CounterAnimal~\cite{wang2024soberlookrobustnessclips}). 
ImageNet-X~\cite{idrissi2022imagenetxunderstandingmodelmistakes} adds 
structured attribute annotations; our reannotation 
extends this idea by combining attribute annotation with multilabel correction 
and localisation. None of these efforts re-examines the inherited label 
structure.



\textbf{\noindent{Large-Scale Annotation Methodology.}} Crowdsourcing platforms such as \ac{amt} and Prolific are the standard infrastructure for large-scale vision benchmarks~\cite{deng2009imagenet, lin2015microsoftcococommonobjects, krishna2016visualgenomeconnectinglanguage}, but they have structural limitations for fine-grained semantic annotation. Guidelines must be fully specified before annotation begins and cannot easily be updated when edge cases arise. Collecting multiple labels per image and resolving disagreements by majority vote reduces random errors but does not fix systematic annotator biases and introduces non-transparent design choices~\cite{lease2011quality}. 
On fine-grained categories, one expert label beats the average of ten crowd labels~\cite{7298658}. Frameworks such as Mephisto~\cite{urbanek2023mephistoframeworkportablereproducible} improve reproducibility but do not address this core problem.

\textbf{\noindent{LLMs as Annotators.}} Whether LLMs can replace or augment human annotators has been actively studied in NLP, where \cite{Gilardi_2023} showed that ChatGPT surpasses crowdworkers on several text classification tasks. In computer vision, prior LLM-based annotation work has focused on image captioning, with quality verified through manual checks~\cite{singla2024pixelsproselargedataset} or downstream performance~\cite{chen2024sharegpt4v}, yet a gap with human annotators remains even for modern LLMs~\cite{OnoeDocci2024}. \cite{kisel2026multimodallargelanguagemodels} provided a case study, showing that GPT-4o \cite{openai2024gpt4ocard} predictions frequently agreed with or improved upon trained human annotations. The complementary strengths of each motivate the combined human–LLM strategy we adopt.

\section{\imnet{} Annotation: From Classification to Multilabel Localization}
\label{sec:task}

Image classification is canonically defined as assigning a single label from a fixed set of categories to an image. 
In practice, however, this single-label assumption is frequently violated:
an image may contain multiple objects, or a single object may admit several valid labels.\footnote{A single object may carry multiple labels simultaneously (\eg a \cls{cardigan} made of \cls{wool}), see Suppl.~\Cref{fig:multi}.}
These issues are particularly pronounced in \imnet{}, which is multilabel \cite{beyer2020imagenet, pmlr-v119-shankar20c, tsipras2020imagenetimageclassificationcontextualizing} (see~\Cref{sec:background}). 
Forcing a single label thus reduces a genuinely multi-valued problem to a selection that is often arbitrary: in \imnet{}, it is typically determined by the search query used to retrieve the image rather than by any property of the image itself, and it often does not correspond to the visually dominant object.
Even adopting a "dominant object" rule would not fully resolve this, since dominance is itself a subjective judgment,
and many images contain multiple objects of comparable prominence.



\begin{figure}[t]
    \centering
    \begin{subfigure}{0.5\linewidth}
        \centering
        \begin{minipage}{0.29\linewidth}
            \includegraphics[width=\linewidth]{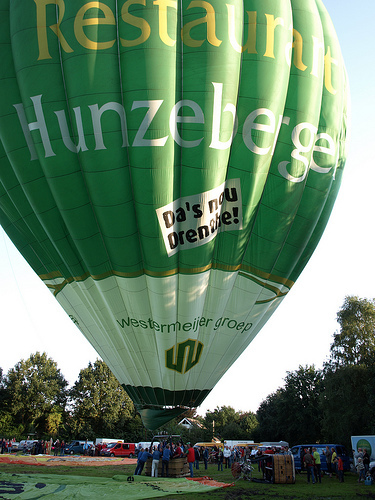}
        \end{minipage}
        \hfill
        \begin{minipage}{0.68\linewidth}
            \includegraphics[width=\linewidth]{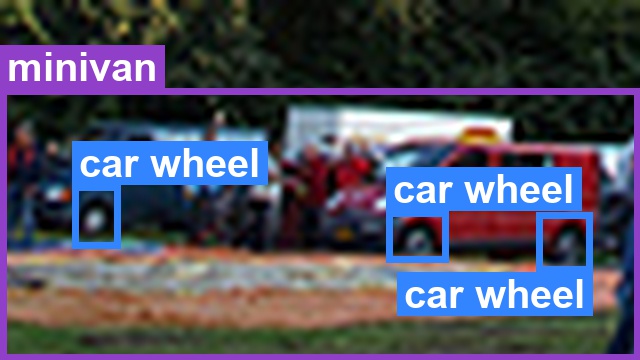}
        \end{minipage}
        \caption{\raggedright
        \gt{}: \textcolor{ForestGreen}{balloon} \hphantom{ }
        \regt{}: \textcolor{ForestGreen}{balloon}; \textcolor{ForestGreen}{minivan}; \textcolor{ForestGreen}{car wheel}
        }
    \end{subfigure}
    \hfill
    \begin{subfigure}{0.46\linewidth}
        \centering
        \begin{minipage}{0.63\linewidth}
            \includegraphics[width=\linewidth]{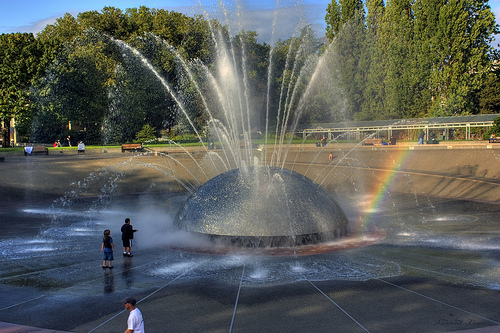}
        \end{minipage}
        \hfill
        \begin{minipage}{0.34\linewidth}
            \includegraphics[width=\linewidth]{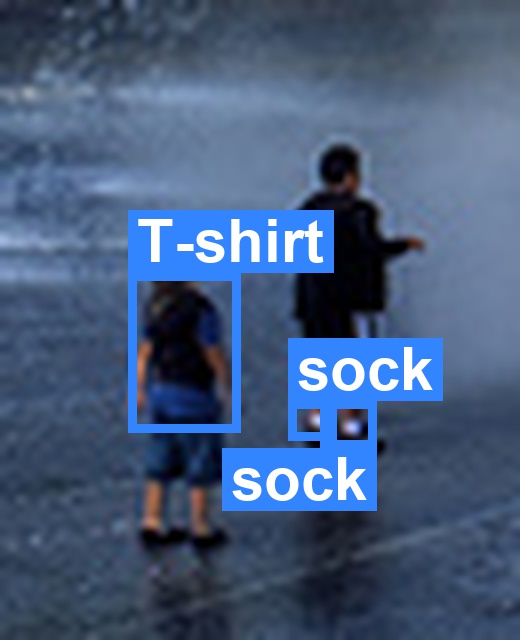}
        \end{minipage}
        \caption{\raggedright
        \gt{}: \textcolor{ForestGreen}{fountain} \hphantom{ }
        \regt{}: \textcolor{ForestGreen}{fountain}; \textcolor{ForestGreen}{T-shirt}; \textcolor{ForestGreen}{sock}
        }
    \end{subfigure}
    \caption{Small-scale objects and their zoomed-in crops (right). 
    Such objects are recognizable only from the context.
    Initially, we aimed to annotate every visible object in 
    the image; small objects, as shown in (a) and (b), right image, discovered during annotation, led to a revision of the definition of what to annotate, see~\Cref{sec:task}.
    The \cls{minivan} bounding box has the \textcolor[HTML]{9040C8}{\rule{1.2ex}{1.2ex}} \attr{crowd} attribute. 
    } \label{fig:size} \end{figure}

We thus redefine the \imnet{} annotation task as a multilabel 
assignment and localization of all objects that belong to one of the 
1{,}000 \imnet{} classes, and at the same time, immediately 
when the image is viewed~(see~~\Cref{fig:size}). 
The latter condition is motivated by the fact that objects spanning only a few pixels are recognisable purely from context rather than appearance, and labelling them makes the label set subjective and dependent on how hard each annotator chooses to look.
Localization is the natural representation for multilabel annotation: it disambiguates which instance carries each label and extends the benchmark utility beyond classification to object detection, counting, and crop-based evaluation protocols, while also enabling deeper analysis of model behavior, explainability, and interpretability.
Each object is annotated with a bounding box enclosing it with a moderate margin.\footnote{Bounding boxes throughout the paper may appear looser than in the actual annotations for illustration purposes, to help the reader distinguish objects in complex scenes.}

Several existing ImageNet-derived benchmarks isolate images with specific visual properties in dedicated datasets~\cite{hendrycks2021many, 
wang2019learningrobustglobalrepresentations, NEURIPS2019_97af07a1, 
moayeri2022hard}. 
We adopt a similar approach, explicitly marking objects carrying unique visual traits in the form of optionally assigned \attr{attributes}. 
Two attributes affect the evaluation 
protocol: \attr{dominant} identifies the object a single-label prediction 
should match, and \attr{crowd} marks groups of instances covered by one box. 
Three isolate specific model capabilities: \attr{rendition} for non-photographic 
depictions, \attr{reflection} for objects visible only on reflective surfaces, 
and \attr{text-recognition} for objects identifiable only by reading text in 
the image~(see~\Cref{fig:attributes} and 
Suppl.~\Cref{fig:rendition,fig:crowd,fig:ocr,fig:reflected,fig:saliency}).
\section{The Reannotation Process}
\label{sec:reannotation}

The original \imnet{} labels are single-label, frequently miss objects that 
are present, and contain classes with overlapping definitions~(\Cref{sec:background}).
This is largely a consequence of relying on minimally trained crowd workers.
We designed a pipeline built around trained in-house annotators, continuous 
communication, and quality control. 
\Cref{subsec:workflow} describes the annotation team and the preparation process each annotator follows before labelling images. 
\Cref{subsec:tools} covers the tools and model-based assistance available. 
\Cref{subsec:iterative} discusses how the pipeline evolved over time. \Cref{subsec:limitations} reflects on its inherent limitations.

\subsection{Annotation Workflow}
\label{subsec:workflow} 

The workflow consists of five components: creating the annotator pool, general 
annotator training to perform the task defined in~\Cref{sec:task}, organising classes into groups and assigning them to 
annotators, per-class preparation, and image-level labelling. Each is 
described below.


\textbf{\noindent{Annotator pool.}} 
The annotation was carried out by a team of 7 non-domain-expert annotators recruited and trained in-house, spanning different age groups (16--50 years old), educational (from high school to university) and cultural backgrounds, geology specialists, canine enthusiasts, and car buffs. 
The number of active annotators varied during the project, but
one lead annotator was present throughout the entire process. 
This role demanded a basic understanding of machine learning concepts, as decisions such as class name revision are driven by their implications for model evaluation.

\textbf{\noindent{General annotator training.}} 
Initial training covered familiarization with the problematic class 
relationships identified in prior work and how to address them, as well 
as approved annotated examples for each \attr{attribute}.
Hands-on practice with a set of intentionally challenging classes chosen by the authors was also included. 
Full details of the training process are provided in Suppl.~\Cref{subsec:supp-training}.

\textbf{\noindent{Class groups.}}
Rather than annotating classes in isolation, annotators were assigned classes in groups constructed by clustering classes that share a common parent node in the WordNet~\cite{miller-1994-wordnet} hierarchy (source for \imnet{} classes). 
Although we cannot know in advance which specific classes co-occur or cause confusion in a given image (both of which happen 
frequently, see~\Cref{sec:background}), grouping classes by their shared parent increases the likelihood that problematic class pairs are presented to annotators together in one batch.
Another motivation for grouping is to reduce the cognitive burden: with 1,000 classes in total, keeping track of all possible labels simultaneously places an unreasonable load on annotators, and grouping reduces this to a manageable subset at a time.
Where possible, group assignments were made to align with annotators' personal familiarity with the subject matter, to 
encourage engagement with the task.
Details on group composition are provided in Suppl.~\Cref{subsec:supp-groups}.


\textbf{\noindent{Per-class preparation protocol.}}
Before annotating any image, annotators were instructed to follow a structured preparation process. 
Each annotator examined the image content of each class in their group, recorded a working definition and, where needed, a new class name (original names frequently misrepresent content) in a shared table, and resolved any ambiguities collectively with the authors before proceeding. Full details are provided in Suppl.~\Cref{subsec:per-class-supp}.
Only after completing this preparation did annotators proceed to image-level labelling, supported by the tools described in~\Cref{subsec:tools}. 

Throughout all components of the workflow, annotators' work was continuously reviewed and individual feedback provided via control sets and a supervised communication channel (see Suppl.~\Cref{subsec:supp-training}).

\subsection{Annotation Tools and Resources}
\label{subsec:tools}
\textbf{\noindent{Model assistance.}} For each image, annotators were provided with top-5 object instance proposals from detector OWLv2~\cite{minderer2024scalingopenvocabularyobjectdetection}, prompted with a list of all ImageNet-1k class names. 
The generated boxes served as optional starting points that could be accepted, edited, or discarded. 
Alongside the image, the top-20 OpenCLIP~\cite{ilharco_gabriel_2021_5143773} predictions were displayed, together with image examples for each predicted class. 
This helped annotators spot objects they might otherwise overlook.

\textbf{\noindent{Additional resources.}} For fine-grained species identification and discovery of visually similar species, annotators consulted \href{https://www.inaturalist.org}{\textcolor{blue}{iNaturalist}}, which provides expert-verified photographs and taxonomic information. 
The complete \href{https://github.com/klarajanouskova/ImageNet/blob/main/classes/problem_groups/clusters_by_category.json}{\textcolor{blue}{list of known problematic class relationships}} was available to all annotators. 
\href{https://vrg.fel.cvut.cz/reimagenet}{\textcolor{blue}{Class names and definitions}} were recorded in a shared table, serving as the reference for all decisions.

\textbf{\noindent{The annotation application}}
supports two complementary views: a grid-based class-level view for quickly browsing multiple images and reviewing class definitions (see~\Cref{subsec:workflow}), and a detailed image-level view for drawing and editing bounding boxes, assigning \attr{attributes}, and referencing model predictions. 
A full description of the application and its workflow is provided in Suppl.~\Cref{subsec:supp-app}.

\subsection{Annotation as an Iterative Process}
\label{subsec:iterative}

\textbf{\noindent{Initial assumptions.}} Reannotating \imnet{} proved 
considerably harder than anticipated. The pipeline was designed on the 
assumption that prior analysis of the dataset's problems~\cite{kisel2024flaws} 
was enough to specify reliable guidelines in advance, and that trained 
in-house annotators with continuous communication would then be sufficient to 
produce a reliable benchmark. Neither assumption held.

\textbf{\noindent{What continuous communication revealed.}} In practice, with each question in the group chat and every assigned control set (see Suppl.~\Cref{subsec:supp-training}), more edge cases were discovered. 
This caused repeated mid-annotation revisions: new attributes were introduced, class names and definitions updated, and our understanding of which objects should be annotated evolved (see~\Cref{fig:size}). 

\textbf{\noindent{The limits of human annotation.}} A case study in~\cite{kisel2026multimodallargelanguagemodels} showed that on a set of challenging images where GPT-4o disagreed with our annotations, annotators confirmed (admitted they were wrong) or integrated the GPT-4o prediction in approximately 50\% of cases. 
While this demonstrates that \acp{mllm} are capable annotators that can outperform trained humans, the other 50\% (where annotators disagreed with the model and were right) shows that \acp{mllm} are far from a reliable replacement.
Two things follow from this: reliable annotation is inherently iterative and cannot be achieved in one shot; and human+\ac{mllm} collaboration outperforms either in isolation.

\textbf{\noindent{Verification phase.}} We therefore conduct a second verification phase in which annotators revisit already-annotated images using the now finalised guidelines. 
Inspired by the case study results, OWLv2 and OpenCLIP predictions are replaced by \acp{mllm} predictions from~\cite{kisel2026multimodallargelanguagemodels} with object localizations generated by SAM3~\cite{carion2026sam3segmentconcepts}, providing annotators with a stronger optional reference.
\footnote{We also present the single-label ILSVRC2012 ImageNet localizations, which were not utilised until this point.} 
This does not imply that the current annotations are unreliable: we consider them to be of good quality, but verification allows us to catch remaining errors and apply \attr{attributes} introduced mid-annotation.
This process is ongoing; only minor refinements are 
expected, and the current results are representative of the final dataset. 
Examples of changes made during this phase are shown in Suppl.~\Cref{fig:verification}.




\subsection{Limitations}
\label{subsec:limitations}

All annotators share a European background (Czechia, Ukraine, Greece, and 
Latvia), which may affect how certain concepts are interpreted. 
Some classes represent objects or practices outside annotators' cultural experience (\eg \cls{abaya}), and while annotators studied such concepts from material available online, cultural nuance may be lost. 
The same applies to fine-grained wildlife: annotators are non-experts and rely on online resources, which may not fully capture species-level distinctions (a task notoriously difficult even for domain experts~\cite{kisel2024flaws}).

Several further sources of noise are inherent to the pipeline. 
The shared class definition table enforces consistency across annotations, but if a class is defined incorrectly, the error propagates to all images that contain this class. 
Although model predictions are anonymised and presented as optional references, they may still nudge annotators towards certain labels.

Finally, our reannotation covers only the validation set; we make no claims about the training set. 
Due to the distribution shift between training and validation sets~\cite{kisel2024flaws}, our class names and definitions may not accurately reflect the training set, because they are grounded in validation set image content.

We make all annotations, class names, and definitions publicly available with an option to contribute corrections and suggestions via \href{https://vrg.fel.cvut.cz/reimagenet}{\textcolor{blue}{the annotation preview tool}}. 
We see the benchmark as something that should evolve iteratively over time, and invite the community to take part in it.

\section{The Dataset Statistics and Properties}
\label{sec:results}

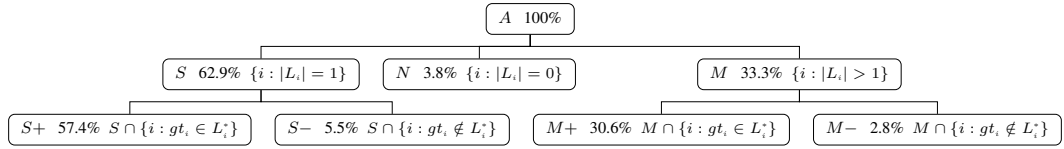
\begin{figure}[ht]
\centering
\begin{adjustbox}{scale=0.77}
\begin{forest}
  for tree={
    draw,
    rounded corners=3pt,
    align=center,
    font={\fontsize{8}{9}\selectfont},
    inner xsep=6pt,
    inner ysep=3pt,
    l sep=4mm,
    s sep=4mm,
    fork sep=2mm,
    edge={-},
  },
  forked edges,
  [{$A$~~~100\%}
    [{$S$~~~62.9\%~~{\scriptsize$\{i:|L_i|=1\}$}}
      [{$S+$~~~57.4\%~~{\scriptsize$S \cap \{i:gt_i \in L_i^*\}$}}]
      [{$S-$~~~5.5\%~~{\scriptsize$S \cap \{i:gt_i \notin L_i^*\}$}}]
    ]
    [{$N$~~~3.8\%~~{\scriptsize$\{i:|L_i|=0\}$}}]
    [{$M$~~~33.3\%~~{\scriptsize$\{i:|L_i|>1\}$}}
      [{$M+$~~~30.6\%~~{\scriptsize$M \cap \{i:gt_i \in L_i^*\}$}}]
      [{$M-$~~~2.8\%~~{\scriptsize$M \cap \{i:gt_i \notin L_i^*\}$}}]
    ]
  ]
\end{forest}
\end{adjustbox}
\caption{Image categorization based on \regt{} and \gt{}.
For image $i$, $gt_i$ is the \gt{} label, $L_i$ the set of \regt{} labels, and $L_i^*$ its extension with semantically equivalent classes (see~\Cref{subsec:eval-protocol}).
$A$ denotes  all,  $S$ single-label, and $M$ multi-label images respectively. Symbols $+$ and $-$ indicate agreement and disagreement, respectively, of \regt{} and \gt{}. 
Images in class $N$ contain no \imnet{} object.
}
\label{fig:reannot_defs}
\end{figure}

\noindent\textbf{Label Statistics.} Following~\cite{kisel2026multimodallargelanguagemodels}, we partition images into categories based on how many \regt{} labels each image has and whether the original \gt{} label is among them, as shown in~\Cref{fig:reannot_defs}.

\begin{wrapfigure}{r}{0.46\textwidth}
    \vspace{-0.75em}
    \centering
    \includegraphics[width=\linewidth]{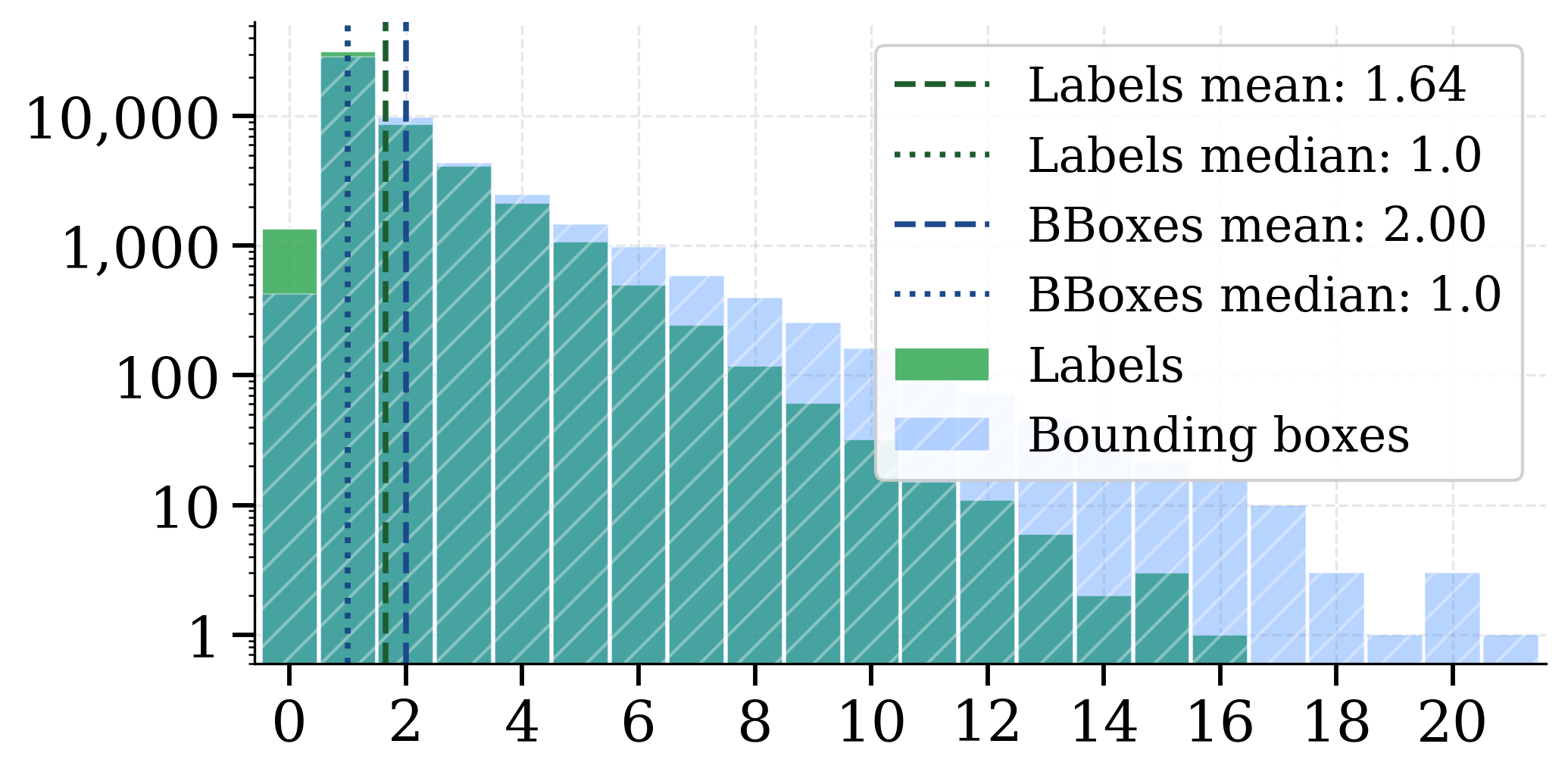}
    \caption{Distributions of the number of labels and bounding boxes per image, 
    log-scaled.}
    \label{fig:dist_reann}
    \vspace{-1.5em}
\end{wrapfigure}


Our multilabel rate of 33.3\% is substantially higher than that of all 
prior reannotation efforts: ImageNet 
ReaL~\cite{beyer2020imagenet}~(14.9\%), Multi-label 
annotations~\cite{pmlr-v119-shankar20c}~(18.3\%), 
ImageNetMultiLabel~\cite{tsipras2020imagenetimageclassificationcontextualizing}~(21.6\%), 
and Label Errors~\cite{northcutt2021pervasivelabelerrorstest}~(11.0\%), 
reflecting our originally exhaustive labeling approach (see~\Cref{fig:size}). Disagreements with labels assigned in prior work are visualized in Suppl.~\Cref{sec:supp_prior_work}.

The distributions of labels and bounding boxes per image are shown in~\Cref{fig:dist_reann}. 
The gap between label and box counts is most commonly due to the appearance of the same class as multiple separate instances in the image, each requiring its own bounding box. 
Additionally, multi-label objects (a single box with multiple labels, Suppl.~\Cref{fig:multi}) and the \attr{crowd} attribute (multiple instances collapsed into one box, ~\Cref{fig:attributes}) influence the statistics.

\subsection{Impact of Labelling on Top-1 Accuracy}
\label{subsec:eval-protocol}



\begin{table*}[bt]
\footnotesize
\centering
\caption{Recognition accuracy $\mathrm{Acc}$ on image categories defined in~\Cref{fig:reannot_defs} for \acp{mllm} (top), \acp{vlm} (middle) and supervised models (bottom). Subscripts denote the number of images in a category. Images in $\mathrm{N}$ contain no valid ImageNet-1k class and are excluded from evaluation. \textcolor{ForestGreen}{Increase} in accuracy on \regt{} with respect to \gt{} is shown for $\mathrm{{A{\setminus}N}_{\cntAnoN}}$.}
\setlength{\tabcolsep}{3pt}
\begin{tabular}{lrlrrrrrrr}
\toprule
& \multicolumn{1}{c}{\gt{}} & & \multicolumn{7}{c}{\regt{}} \\ \cmidrule{2-2} \cmidrule{4-10} 
& \multicolumn{1}{c}{$\mathrm{{A{\setminus}N}_{\cntAnoN}}$} & & \multicolumn{1}{c}{$\mathrm{{A{\setminus}N}_{\cntAnoN}}$} & \multicolumn{1}{c}{$\mathrm{S_{\cntReGTS}}$} & \multicolumn{1}{c}{$\mathrm{{S+}_{\cntReGTSpos}}$} & \multicolumn{1}{c}{$\mathrm{{S-}_{\cntReGTSneg}}$} & \multicolumn{1}{c}{$\mathrm{M_{\cntReGTM}}$} & \multicolumn{1}{c}{$\mathrm{{M+}_{\cntReGTMpos}}$} & \multicolumn{1}{c}{$\mathrm{{M-}_{\cntReGTMneg}}$} \\
\cmidrule{1-10}
\cgpt{} \cite{openai2026gpt54} & 72.54 & & 77.64 {\scriptsize{\textcolor{ForestGreen}{+5.10}}} & 77.86 & 80.92 & 46.11 & 77.22 & 80.59 & 40.16 \\
\qwen{} \cite{yang2025qwen3technicalreport} & 70.95 & & 77.04 {\scriptsize{\textcolor{ForestGreen}{+6.09}}} & 75.97 & 78.88 & 45.71 & 79.08 & 82.26 & 44.05 \\
\third{} \cite{gemma4_2026} & 69.48 & & 74.87 {\scriptsize{\textcolor{ForestGreen}{+5.39}}} & 74.48 & 77.23 & 45.93 & 75.60 & 78.63 & 42.25 \\
\cmidrule{1-10}
\siglip{} \cite{zhai2023sigmoidlosslanguageimage} & 84.87 & & 87.86 {\scriptsize{\textcolor{ForestGreen}{+2.99}}} & 88.54 & 92.86 & 43.68 & 86.60 & 91.09 & 37.06 \\
\sigliptwo{} \cite{tschannen2025siglip2multilingualvisionlanguage} & 85.82 & & 88.43 {\scriptsize{\textcolor{ForestGreen}{+2.61}}} & 89.11 & 93.55 & 42.96 & 87.16 & 91.75 & 36.63 \\
\sigliptwog{} \cite{tschannen2025siglip2multilingualvisionlanguage} & 86.74 & & 88.99 {\scriptsize{\textcolor{ForestGreen}{+2.25}}} & 89.69 & 94.23 & 42.60 & 87.66 & 92.42 & 35.26 \\
\cmidrule{1-10}
\dino{} \cite{simeoni2025dinov3} & 86.49 & & 87.69 {\scriptsize{\textcolor{ForestGreen}{+1.20}}} & 88.97 & 94.08 & 35.87 & 85.28 & 90.29 & 30.06 \\
\effnetv{} \cite{tan2021efficientnetv2smallermodelsfaster} & 87.07 & & 88.23 {\scriptsize{\textcolor{ForestGreen}{+1.16}}} & 88.78 & 93.79 & 36.74 & 87.18 & 92.49 & 28.70 \\
\effnetl{} \cite{xie2020selftrainingnoisystudentimproves} & 89.82 & & 90.52 {\scriptsize{\textcolor{ForestGreen}{+0.70}}} & 90.76 & 96.00 & 36.30 & 90.08 & 95.37 & 31.87 \\
\eva{} \cite{Fang_2024} & 91.33 & & 91.38 {\scriptsize{\textcolor{ForestGreen}{+0.05}}} & 91.52 & 96.91 & 35.58 & 91.11 & 96.68 & 29.78 \\
\bottomrule
\end{tabular}
\label{tab:main_res}
\end{table*}

We measured the top-1 ReaL accuracy~\cite{beyer2020imagenet} of \ac{mllm}, \ac{vlm} and supervised models listed in~\Cref{tab:main_res} 
adapted as in ~\cite{kisel2026multimodallargelanguagemodels} to reflect equivalent class pairs (\eg \cls{laptop} = \cls{notebook computer}).%
\footnote{The full list of equivalent class pairs is available on the \href{https://huggingface.co/datasets/c1rcuslegend/ReImageNet}{\textcolor{blue}{dataset page}}.}
Model prediction $p_i$ is correct, \ie $\alpha_i=1$, if it is in the image $i$ \regt{} label set $L^*_i$, which is the set of labels $L_i$ of image $i$ extended by all equivalent labels from the equivalence set $\mathcal{E}$.
Images in N contain no valid label and are excluded from evaluation. 
Model accuracy $\mathrm{Acc}$ on images set $I$ is then:


\begin{equation}
\mathrm{Acc} = \frac{1}{|I \setminus N|}\sum_{i \in I \setminus N} \alpha_i,
\qquad
\alpha_i = \bigl[ p_i \in L_i^* \bigr]
\qquad
L_i^* = L_i \cup \{b \mid \exists\, a \in L_i,\, \{a, b\} \in \mathcal{E}\}
\end{equation}

Model accuracies are presented in~\Cref{tab:main_res}. All models benefit from reannotation, with \regt{} accuracy consistently higher than \gt{} for all model families (detailed model specifications are provided in Suppl.~\Cref{sec:supp_setup}). 
\acp{mllm} gain the most (GPT-5.4: $+5.10$\%, Qwen3-VL: $+6.09$\%, 
Gemma~4: $+5.39$\%), while gains for supervised models are marginal, nearly vanishing for the strongest model (EffNet-L2: +0.70\%, EVA-02: +0.05\%). This confirms that models less reliant on supervised training signals are more sensitive to label 
quality. 
On the challenging $S-$ and $M-$ subsets, where \gt{} disagrees with \regt{}, all models have accuracy in the 29--46\% range, confirming that these images contain difficult cases.
These findings are consistent with~\cite{kisel2026multimodallargelanguagemodels} on a partial reannotation, now confirmed on the complete validation set.


\begin{table*}[ht]
\footnotesize
\centering
\caption{Accuracy $\mathrm{Acc}$ (with \regt{}) for images in the $\mathrm{S}$ category per visual \attr{attribute}. Subscripts denote the number of images in a subset.
\textcolor{ForestGreen}{Increase} or \textcolor{red}{decrease} in accuracy with respect to $\mathrm{S}$ are shown. $\mathit{Mix}$ and $\emptyset$ denote 
images containing multiple attributes, and no attributes, respectively.}
\setlength{\tabcolsep}{4pt} 

\begin{tabular}{l l llllll}
\toprule
Model & $\mathrm{S}_{\cntReGTS}$ & $\mathit{Crowd}_{1931}$ & $\mathit{Refl}_{228}$ &  $\mathit{Rend}_{410}$ & $\mathit{Text}_{284}$ & $\mathit{Mix}_{88}$ & $\mathit{\emptyset}_{28507}$ \\
\midrule
\cgpt{} & 77.86 &
77.94 {\scriptsize{\textcolor{ForestGreen}{+0.08}}} &

75.44 {\scriptsize{\textcolor{red}{-2.42}}} &
75.61 {\scriptsize{\textcolor{red}{-2.25}}} &
87.32 {\scriptsize{\textcolor{ForestGreen}{+9.46}}} &
71.59 {\scriptsize{\textcolor{red}{-6.27}}} &
77.83 {\scriptsize{\textcolor{red}{-0.03}}} \\
\qwen{} & 75.97 &
78.72 {\scriptsize{\textcolor{ForestGreen}{+2.75}}} &

77.19 {\scriptsize{\textcolor{ForestGreen}{+1.22}}} &
74.63 {\scriptsize{\textcolor{red}{-1.34}}} &
91.55 {\scriptsize{\textcolor{ForestGreen}{+15.58}}} &
79.55 {\scriptsize{\textcolor{ForestGreen}{+3.58}}} &
75.62 {\scriptsize{\textcolor{red}{-0.35}}} \\
\third{} & 74.48 &
77.63 {\scriptsize{\textcolor{ForestGreen}{+3.15}}} &

74.12 {\scriptsize{\textcolor{red}{-0.36}}} &
72.93 {\scriptsize{\textcolor{red}{-1.55}}} &
87.32 {\scriptsize{\textcolor{ForestGreen}{+12.84}}} &
67.05 {\scriptsize{\textcolor{red}{-7.43}}} &
74.18 {\scriptsize{\textcolor{red}{-0.30}}} \\
\midrule
\siglip{} & 88.54 &
89.95 {\scriptsize{\textcolor{ForestGreen}{+1.41}}} &

90.35 {\scriptsize{\textcolor{ForestGreen}{+1.81}}} &
84.88 {\scriptsize{\textcolor{red}{-3.66}}} &
95.42 {\scriptsize{\textcolor{ForestGreen}{+6.88}}} &
90.91 {\scriptsize{\textcolor{ForestGreen}{+2.37}}} &
88.40 {\scriptsize{\textcolor{red}{-0.14}}} \\
\sigliptwo{} & 89.11 &
89.59 {\scriptsize{\textcolor{ForestGreen}{+0.48}}} &

91.23 {\scriptsize{\textcolor{ForestGreen}{+2.12}}} &
84.88 {\scriptsize{\textcolor{red}{-4.23}}} &
96.48 {\scriptsize{\textcolor{ForestGreen}{+7.37}}} &
89.77 {\scriptsize{\textcolor{ForestGreen}{+0.66}}} &
89.04 {\scriptsize{\textcolor{red}{-0.07}}} \\
\sigliptwog{} & 89.69 &
89.23 {\scriptsize{\textcolor{red}{-0.46}}} &

93.42 {\scriptsize{\textcolor{ForestGreen}{+3.73}}} &
85.85 {\scriptsize{\textcolor{red}{-3.84}}} &
96.13 {\scriptsize{\textcolor{ForestGreen}{+6.44}}} &
94.32 {\scriptsize{\textcolor{ForestGreen}{+4.63}}} &
89.67 {\scriptsize{\textcolor{red}{-0.02}}} \\
\midrule
\dino{} & 88.97 &
91.51 {\scriptsize{\textcolor{ForestGreen}{+2.54}}} &

89.47 {\scriptsize{\textcolor{ForestGreen}{+0.50}}} &
81.71 {\scriptsize{\textcolor{red}{-7.26}}} &
90.14 {\scriptsize{\textcolor{ForestGreen}{+1.17}}} &
90.91 {\scriptsize{\textcolor{ForestGreen}{+1.94}}} &
88.88 {\scriptsize{\textcolor{red}{-0.09}}} \\
\effnetv{} & 88.78 &
90.11 {\scriptsize{\textcolor{ForestGreen}{+1.33}}} &

91.23 {\scriptsize{\textcolor{ForestGreen}{+2.45}}} &
79.02 {\scriptsize{\textcolor{red}{-9.76}}} &
82.04 {\scriptsize{\textcolor{red}{-6.74}}} &
86.36 {\scriptsize{\textcolor{red}{-2.42}}} &
88.88 {\scriptsize{\textcolor{ForestGreen}{+0.10}}} \\
\effnetl{} & 90.76 &
92.96 {\scriptsize{\textcolor{ForestGreen}{+2.20}}} &

90.35 {\scriptsize{\textcolor{red}{-0.41}}} &
84.88 {\scriptsize{\textcolor{red}{-5.88}}} &
86.27 {\scriptsize{\textcolor{red}{-4.49}}} &
89.77 {\scriptsize{\textcolor{red}{-0.99}}} &
90.75 {\scriptsize{\textcolor{red}{-0.01}}} \\
\eva{} & 91.52 &
94.10 {\scriptsize{\textcolor{ForestGreen}{+2.58}}} &

92.98 {\scriptsize{\textcolor{ForestGreen}{+1.46}}} &
87.32 {\scriptsize{\textcolor{red}{-4.20}}} &
95.42 {\scriptsize{\textcolor{ForestGreen}{+3.90}}} &
95.45 {\scriptsize{\textcolor{ForestGreen}{+3.93}}} &
91.35 {\scriptsize{\textcolor{red}{-0.17}}} \\
\bottomrule
\end{tabular}
\label{tab:attribute_acc}
\end{table*}

\noindent\textbf{Attribute analysis.} 
\Cref{tab:attribute_acc} breaks down \regt{} accuracy on single-label images ($S$) by \attr{attribute}. The \attr{crowd} attribute improves accuracy for most models (up to +3\%).
Reflection shows a positive effect, strongest for \acp{vlm} (+1.8 to +3.7\%), while GPT-5.4, Gemma 4, and EffNet-L2 show small drops.

The \attr{rendition} attribute degrades performance for all models, as stylized depictions of objects lie on the long tail of the visual distribution. 
The drop is smallest for \acp{mllm} ($-1.3$\% to $-2.3$\%) and most severe for supervised models (EffNetV2: $-9.8$\%, DINOv3: $-7.3$\%), suggesting that language supervision provides stronger generalisation to stylized depictions than standard supervised training.

The \attr{text-recognition} 
attribute shows clear separation across model families: 
\acp{mllm} and \acp{vlm} gain substantially ($+6$--$16$\%), as language-vision alignment lets them leverage visible text as a semantic feature, while standard supervised models 
without such alignment suffer drops (\eg EffNetV2: $-6.7$\%). EVA-02
and \dino{} also benefit, unlike other 
supervised models. \qwen{} achieves the largest gain 
($+15.6$\%), consistent with its results on text-understanding 
benchmarks~\cite{wang2025mathcodervl, wang2024measuring, 
fu2025ocrbenchv2improvedbenchmark, masry2022chartqabenchmarkquestionanswering}.

\subsection{Impact of Localized Annotations}
\label{subsec:crops}

We create a tight crop 
for each bounding box, 
yielding 96{,}051 crops with an average area of 33\% of the 
original image. 
We evaluate them in four setups. 
\textbf{All} treats each crop as an independent sample. 
\textbf{Largest} evaluates
one crop per image, corresponding to the largest bounding box, 
approximating the classical ImageNet center-crop evaluation.
\textbf{Any} considers an image correct if the model correctly classifies at least one of its crops, 
serving as an upper bound on object-centric accuracy. 

\begin{table*}[ht]
\footnotesize
\begin{minipage}[t]{0.565\textwidth}
\centering
\caption{$\mathrm{Acc}$ on object crops in evaluation 
settings described in~\Cref{subsec:crops}. Deltas indicate 
\textcolor{ForestGreen}{increase}/\textcolor{red}{decrease} w.r.t. full-image accuracy presented in the Image column.}
\setlength{\tabcolsep}{2.5pt}
{
\begin{tabular}{l l l lll l l l l}
\toprule
\# Imgs. $\rightarrow$& \multicolumn{3}{c}{$50{,}000$} & & $96{,}051$ & & $38{,}628$ \\
\cmidrule{2-4} \cmidrule{6-6} \cmidrule{8-8}
Model $\downarrow$ & \multicolumn{1}{c}{Image} & \multicolumn{1}{c}{Largest} & \multicolumn{1}{c}{Any} & & \multicolumn{1}{c}{All} & & \multicolumn{1}{c}{ExCrops} \\
\cmidrule{1-8}
\third{} & 75.41 & 69.76 {\tiny\textcolor{red}{-5.6}} & 81.69 {\tiny\textcolor{ForestGreen}{+6.3}} & & 59.68 {\tiny\textcolor{red}{-15.7}} & & 65.78 {\tiny\textcolor{red}{-9.6}} \\
\cmidrule{1-8}
\siglip{} & 87.99 & 81.26 {\tiny\textcolor{red}{-6.7}}& 91.87 {\tiny\textcolor{ForestGreen}{+3.9}}  & & 69.13 {\tiny\textcolor{red}{-18.9}} & & 77.34 {\tiny\textcolor{red}{-10.7}} \\
\sigliptwo{} & 88.54 & 81.45 {\tiny\textcolor{red}{-7.1}} & 92.19 {\tiny\textcolor{ForestGreen}{+3.7}}  & & 69.24 {\tiny\textcolor{red}{-19.3}}& & 77.95 {\tiny\textcolor{red}{-10.6}} \\
\sigliptwog{} & 89.00 & 82.21 {\tiny\textcolor{red}{-6.8}}& 92.96 {\tiny\textcolor{ForestGreen}{+4.0}}  & & 71.41 {\tiny\textcolor{red}{-17.6}} & & 78.87 {\tiny\textcolor{red}{-10.1}} \\
\cmidrule{1-8}
\dino{} & 87.75 & 78.46 {\tiny\textcolor{red}{-9.3}}&  89.75 {\tiny\textcolor{ForestGreen}{+2.0}}  & & 55.44 {\tiny\textcolor{red}{-32.3}} & & 74.41 {\tiny\textcolor{red}{-13.3}} \\
\effnetv{} & 88.31 & 79.57 {\tiny\textcolor{red}{-8.7}}& 90.90 {\tiny\textcolor{ForestGreen}{+2.6}}  & & 63.74 {\tiny\textcolor{red}{-24.6}} & & 76.67 {\tiny\textcolor{red}{-11.6}} \\
\effnetl{} & 90.53 & 79.49 {\tiny\textcolor{red}{-11.0}}& 90.74 {\tiny\textcolor{ForestGreen}{+0.2}}  & & 59.12 {\tiny\textcolor{red}{-31.4}} & & 77.28 {\tiny\textcolor{red}{-13.3}} \\
\eva{} & 91.43 & 80.64 {\tiny\textcolor{red}{-10.8}}& 92.16 {\tiny\textcolor{ForestGreen}{+0.7}}  & & 57.95 {\tiny\textcolor{red}{-33.5}} & & 77.81 {\tiny\textcolor{red}{-13.6}} \\
\bottomrule
\end{tabular}
}
\label{tab:crop_results}
\end{minipage}
\hfill
\begin{minipage}[t]{0.415\textwidth}
\small
\centering
\caption{Lower bounds on annotation error in derivative benchmarks. Computed by aggregating predictions from \cgpt{} and \sigliptwog{}.
Datasets -- \\ IN-x:
ImageNet-V2, -A, -R, 
-Sketch, ObjN: ObjectNet, CoAN: CounterAnimal.
}
\label{tab:error_estimation}
\setlength{\tabcolsep}{2pt}
\begin{tabular}{lrrrrr}
    \toprule
     Data & \# Imgs. & $\hat{\varepsilon}_{\text{IN1k}}$ & $\hat{\varepsilon}_{\text{bench}}$ & $|\mathcal{I}_{\text{bench}}|$ & $\hat{N}_{\text{err}}$ \\
    \midrule
    INV2       & 10{,}000 & 4.74 & 11.15 & 6{,}801 & 436 \\
    IN-A       & 7{,}500  & 2.57 & 14.80 & 3{,}541 & 433 \\
    IN-R       & 30{,}000 & 2.97 & 4.99 & 22{,}622 & 456 \\
    IN-S  & 50{,}889 & 4.74 & 13.38 & 33{,}120 & 2{,}861 \\  
    ObjN        & 18{,}574 & 5.30 & 8.96 & 11{,}916 & 436 \\  
    CoAn   & 13{,}334 & 1.95 & 4.76 & 10{,}444 & 293 \\
    \bottomrule
\end{tabular}
\end{minipage}
\end{table*}

In \textbf{ExCrops}, each tight crop is expanded maximally in all four 
directions without overlapping any other bounding box, computed via an 
R-tree algorithm. Boxes that overlap with another box before applying this procedure are excluded. 
The resulting 38{,}628 ExCrops (average area of 82\%) are guaranteed to contain exactly one annotated object with no overlap with other annotations. ExCrops are backward-compatible with 
classical single-label object-centric evaluation (see Suppl.~\Cref{tab:s_versus_dom}).

Results are shown in~\Cref{tab:crop_results}. The gap between Any and 
Largest shows that in $\sim$10\% of cases models fail to correctly classify the largest object in the image, yet succeed on a smaller or less salient one. 
This questions the standard ImageNet evaluation preprocessing.
The drop on All crops is most severe for supervised models 
(EVA-02: $-33.5$\%, DINOv3: $-32.3$\%) and notably smaller for 
\acp{mllm}. This is likely because supervised ImageNet models are trained on full 
images and are not optimized for small, tightly cropped objects (see~\Cref{fig:size}), unlike 
\acp{mllm} trained on more diverse data. ExCrops recover $\sim$10\% over All crops, confirming that additional context around each object helps.




\subsection{Impact of Errors on Derived Benchmarks}
\label{subsec:derived}
The lower bound on the error rate of derivative benchmarks is 
estimated in~\Cref{tab:error_estimation}. We select two web-scale 
models with uncorrelated error patterns~\cite{kisel2026multimodallargelanguagemodels}, 
\cgpt{} and \sigliptwog{}. 
We then construct an intersection set $\mathcal{I}_{\text{IN1k}}$ of images where both 
models agree on the predicted label. Images with no valid ReGT label are excluded from $\mathcal{I}_{\text{IN1k}}$, consistent with \Cref{subsec:eval-protocol}.
Evaluating $\mathcal{I}_{\text{IN1k}}$ against \regt{} yields an estimated 
error rate $\hat{\varepsilon}_{\text{IN1k}}$. 
We apply the same procedure to each derivative set  restricting 
$\mathcal{I}_{\text{IN1k}}$ to classes present in the target 
benchmark to obtain $\mathcal{I}_{\text{bench}}$. 
The resulting $\hat{\varepsilon}_{\text{bench}}$ are 1.7--5.8$\times$ higher than 
$\hat{\varepsilon}_{\text{IN1k}}$, indicating that derivative sets 
inherit and amplify labelling errors. 
Assuming consistent model behaviour on \imnet{} and its derivatives, 
$(\hat{\varepsilon}_{\text{bench}} - 
\hat{\varepsilon}_{\text{IN1k}}) \times 
|\mathcal{I}_{\text{bench}}|$ provides a lower bound on the number 
of mislabelled images $\hat{N}_{\text{err}}$ in each benchmark. 

\section{Conclusion}
\label{sec:conclusion}

We presented a from-scratch reannotation of the \imnet{} validation set. The original
single-label task is well-defined only on a subset of images; elsewhere we recast it as
multilabel object localisation with revised class definitions.
Of the 50{,}000 images, $\approx$12\% have incorrect \imnet{} labels, 33.3\% are multilabel
and 3.8\% contain no valid \imnet{} class.
\acp{mllm} benefit most from reannotation ($+5$--$6$\% vs.\ 
up to $1.2$\% for supervised models). Attribute-level analysis reveals that 
language-vision alignment lets \acp{mllm} leverage visible text 
($+9$--$16$\%). All model families struggle with stylised 
depictions (\ie \attr{renditions}).
Object-centric evaluation on bounding-box crops shows 
that supervised models lose up to 33\% accuracy when scene context 
is removed, whereas \acp{mllm} are substantially more robust. 
Annotation errors propagate into derivative benchmarks at 
1.7--5.8$\times$ the \imnet{} rate.
Our annotation experience confirms that without substantial training or
domain expertise, an LLM alone outperforms crowd annotators; human--LLM
collaboration outperforms both.

\paragraph{Implications and recommendations.}
Errors in the original benchmark propagate into its derivative evaluation sets
(\Cref{subsec:derived}), and the same structural pressures now reproduce themselves in
modern \ac{mllm} benchmarks: whether crowdsourced under the conditions that produced
\imnet{}'s noise~\cite{chen2024we, meng2024mmiumultimodalmultiimageunderstanding}, or
annotated by the very models they are meant to
evaluate~\cite{ding2024unleashing, liu2023visualinstructiontuning}, which risks measuring
consistency with the annotator-model's training distribution rather than genuine capability.
The lessons of \imnet{} have, by and large, not been absorbed; the cost of ignoring them
keeps rising. Noisy benchmarks were defensible when models trailed humans by a wide
margin. Today, \acp{mllm} routinely match or exceed untrained crowd
annotators on the labelling task itself.

We argue in favour of abandoning the one-shot model of benchmark construction.
The highest quality currently achievable at scale comes from trained annotators paired
with anonymised \ac{mllm} predictions and dedicated tooling, with a lead annotator
bridging the research and labelling teams. 
Where domain-engaged resources exist, whether citizen-science platforms (iNaturalist,
eBird) or expert-curated databases (mycological references and similar), they provide a
level of curation that generic crowdsourcing cannot match and should serve as anchors for
fine-grained class annotations.


\paragraph{Limitations and Broader Impacts.}
\label{subsec:limitations-impact}

The reannotation retains residual inter-annotator variance: which objects to notice, which to mark as dominant, and where exactly to
draw bounding boxes. The pipeline relies on closed-source \acp{mllm} (GPT-4o,
GPT-5.4) whose outputs are not repeatable and whose availability may change.
More accurate \imnet{} ground truth and a transferable methodology for estimating label
noise enable better-calibrated claims about model performance and lower the cost of
auditing widely-used benchmarks; on the other hand, (i) our work inherits \imnet{}'s
known issues around problematic categories and image provenance, which reannotation does
not address;
(ii) defining label quality via model-ensemble agreement risks entrenching current
architectures' inductive biases; and
(iii) a "cleaner" benchmark may produce unwarranted confidence in evaluation results that
remain limited by distribution shift and construction validity.

\clearpage

\bibliographystyle{abbrvnat}
\bibliography{references}

\clearpage

\appendix

\section{Annotation Pipeline}
\label{sec:supp_annotation_guidelines}

\begin{figure}[b]
    \centering
    \begin{subfigure}[t]{0.23\linewidth}
        \vspace{0pt}
        \centering
        \includegraphics[width=\linewidth]{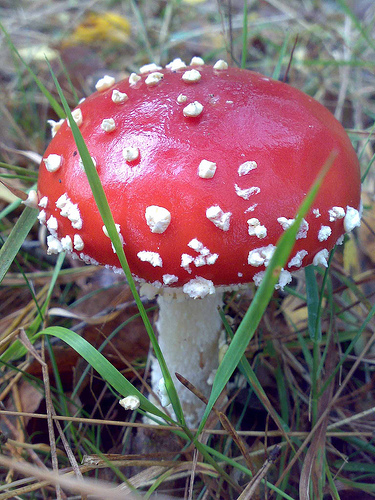}
        \caption{
        \gt{}: \textcolor{ForestGreen}{agaric} \\ 
        \hphantom{(a) }\regt{}: \textcolor{ForestGreen}{agaric}{; } \\
        \hphantom{(a) ReGT: }\textcolor{ForestGreen}{mushroom}
        }
    \end{subfigure}
    \hfill
    \begin{subfigure}[t]{0.23\linewidth}
        \vspace{0pt}
        \centering
        \includegraphics[width=\linewidth]{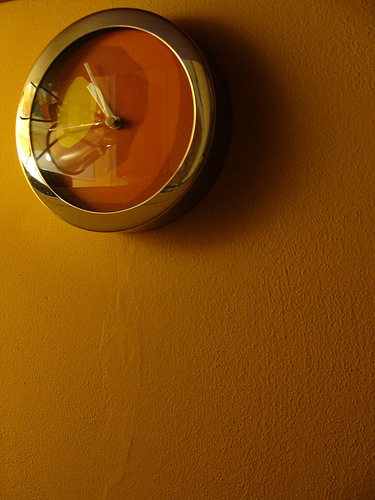}
        \caption{
        \gt{}: \textcolor{ForestGreen}{wall clock} \\ 
        \hphantom{(a) }\regt{}: \textcolor{ForestGreen}{wall clock}{; } \\
        \hphantom{(a) ReGT: }\textcolor{ForestGreen}{analog clock}
        }
    \end{subfigure}
    \hfill
    \begin{subfigure}[t]{0.23\linewidth}
        \vspace{0pt}
        \centering
        \includegraphics[width=\linewidth]{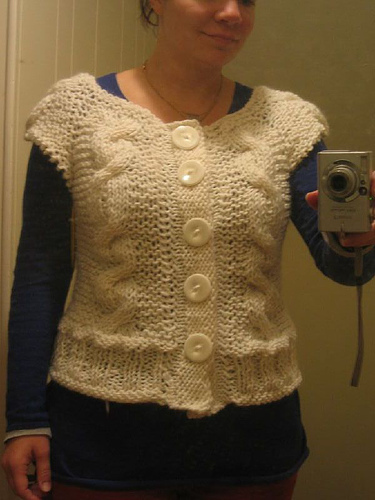}
        \caption{
        \gt{}: \textcolor{ForestGreen}{cardigan} \\ 
        \hphantom{(a) }\regt{}: \textcolor{ForestGreen}{cardigan; wool}
        }
    \end{subfigure}
    \hfill
    \begin{subfigure}[t]{0.245\linewidth}
        \vspace{0pt}
        \centering
        \includegraphics[width=\linewidth]{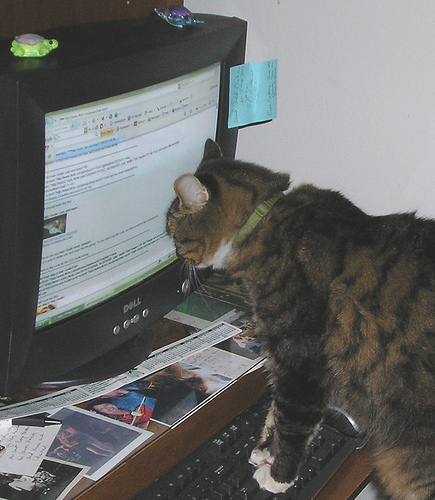}
        \caption{
        \gt{}: \textcolor{ForestGreen}{CRT screen} \\ 
        \hphantom{(a) }\regt{}: \underline{\textcolor{ForestGreen}{CRT screen}}{; } \\
        \textcolor{ForestGreen}{computer keyboard}{; }\textcolor{ForestGreen}{website}{; }\underline{\textcolor{ForestGreen}{monitor}}
        }
    \end{subfigure}
    \caption{Multilabel objects in \imnet{}, \ie objects with multiple valid labels. 
    Some objects exhibit strict hierarchy, where a class is a subclass of another: \eg (a) all \cls{agaric} are \cls{mushroom}. 
    Super-classes are added automatically, so annotators only need to label the most specific class. 
    Other classes show more complex relations: \eg (b) this \cls{wall clock} is also an \cls{analog clock}, but not all wall clocks are analog, (c) this specific \cls{cardigan} is made of \cls{wool}, but not all cardigans are, (d) this \cls{CRT screen} is a \cls{monitor}, but not all CRTs are. 
    In these cases, the annotators need to label all of the classes.}
    \label{fig:multi}
\end{figure}
\begin{figure}[t]
    \centering
    \begin{subfigure}[t]{0.21\linewidth}
        \vspace{0pt}
        \centering
        \includegraphics[width=\linewidth]{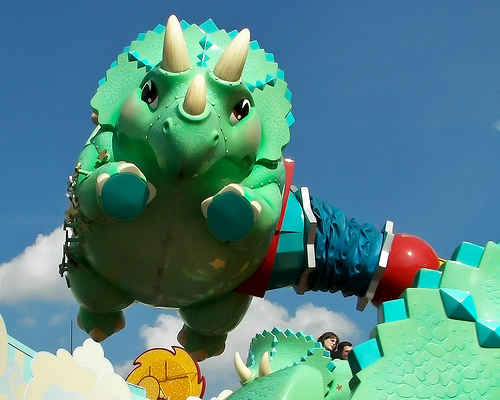}
        \caption{
        \gt{}: \textcolor{ForestGreen}{triceratops} \\ 
        \hphantom{(a) }\regt{}:  \underline{\textcolor{ForestGreen}{triceratops}}
        }
    \end{subfigure}
    \hfill
    \begin{subfigure}[t]{0.23\linewidth}
        \vspace{0pt}
        \centering
        \includegraphics[width=\linewidth]{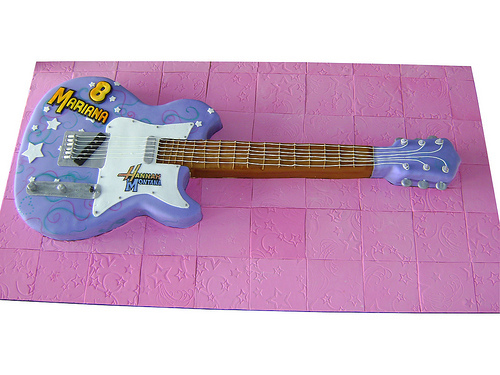}
        \caption{
        \gt{}:  \textcolor{ForestGreen}{electric guitar} \\ 
        \hphantom{(a) }\regt{}: \underline{\textcolor{ForestGreen}{electric guitar}}
        }
    \end{subfigure}
    \hfill
    \begin{subfigure}[t]{0.24\linewidth}
        \vspace{0pt}
        \centering
        \includegraphics[width=\linewidth]{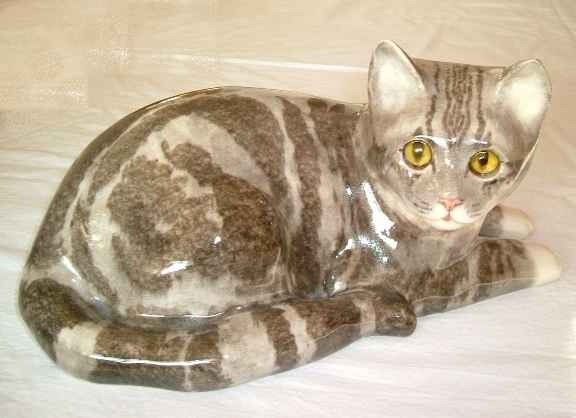}
        \caption{
        \gt{}:  \textcolor{ForestGreen}{tabby cat} \\ 
        \hphantom{(a) }\regt{}:  \underline{\textcolor{ForestGreen}{tabby cat}}
        }
    \end{subfigure}
    \hfill
    \begin{subfigure}[t]{0.24\linewidth}
        \vspace{0pt}
        \centering
        \includegraphics[width=\linewidth]{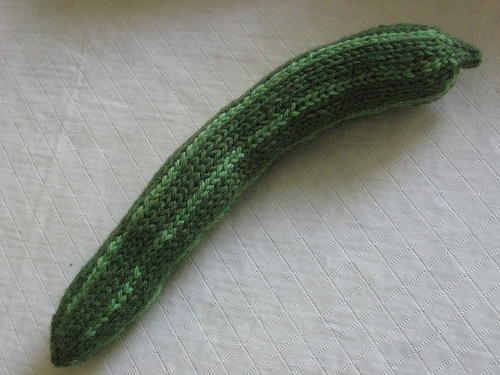}
        \caption{
        \gt{}:  \textcolor{ForestGreen}{cucumber} \\ 
        \hphantom{(a) }\regt{}:  \underline{\textcolor{ForestGreen}{cucumber}}{; }\textcolor{ForestGreen}{wool}
        }
    \end{subfigure}

    \caption{Images containing objects with the \underline{\attr{rendition}} attribute. The attribute is used when the depicted object visually mimics an instance of the original class, but appears as a toy, drawing, sculpture, or other representation. 
    In some categories (\eg \cls{triceratops}), real photographs cannot exist, and all images are renditions. In~(d), \cls{cucumber} is a multilabel object, see~\Cref{fig:multi}.}
    \label{fig:rendition}
\end{figure}

\begin{figure}[t]
    \centering
    \begin{subfigure}[t]{0.32\linewidth}
        \vspace{0pt}
        \centering
        \includegraphics[width=\linewidth]{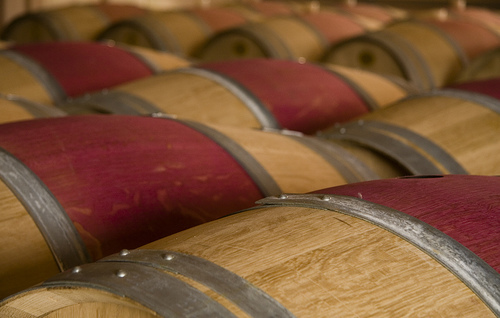}
        \caption{\raggedright
        \gt{}: \textcolor{ForestGreen}{barrel} \\ 
        \hphantom{(a) }\regt{}: \underline{\textcolor{ForestGreen}{barrel}}
        }
    \end{subfigure}
    \hfill
    \begin{subfigure}[t]{0.31\linewidth}
        \vspace{0pt}
        \centering
        \includegraphics[width=\linewidth]{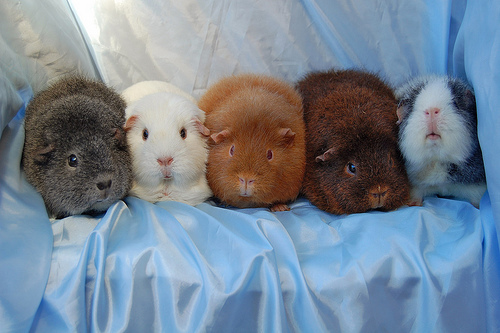}
        \caption{
        \gt{}:  \textcolor{ForestGreen}{guinea pig} \\ 
        \hphantom{(a) }\regt{}:  \underline{\textcolor{ForestGreen}{guinea pig}}
        }
    \end{subfigure}
    \hfill
    \begin{subfigure}[t]{0.335\linewidth}
        \vspace{0pt}
        \centering
        \includegraphics[width=\linewidth]{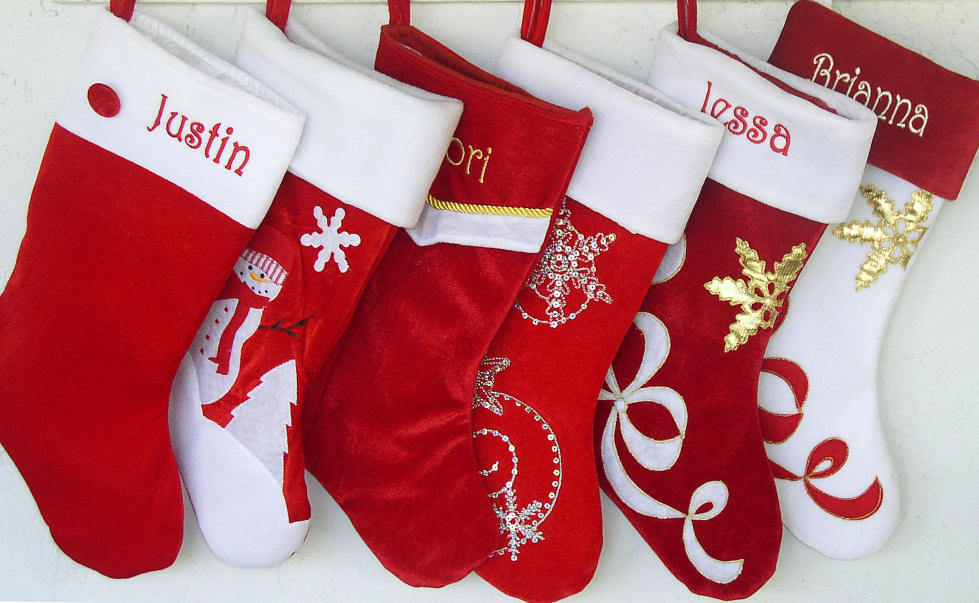}
        \caption{
        \gt{}:  \textcolor{ForestGreen}{Christmas stocking} \\ 
        \hphantom{(a) }\regt{}:  \underline{\textcolor{ForestGreen}{Christmas stocking}}
        }
    \end{subfigure}

    \caption{Images containing bounding boxes with the \underline{\attr{crowd}} attribute. The attribute is used when the image depicts five or more instances of the same class, all of which share identical values for the other (optional) attributes, to simplify the annotation process.}
    \label{fig:crowd}
\end{figure}

\begin{figure}[t]
    \centering
    \begin{subfigure}{0.27\linewidth}
        \centering
        \includegraphics[width=\linewidth]{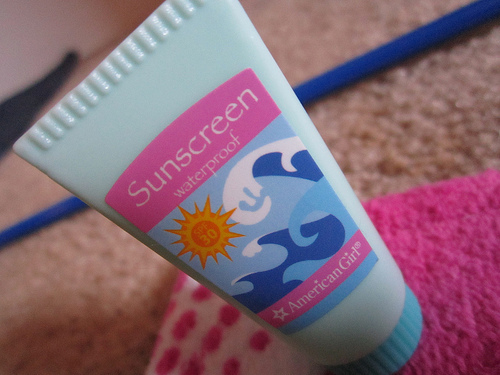}
        \caption{
        \gt{}$=$\regt{}: \\ 
        \hphantom{(a) }\textcolor{ForestGreen}{sunscreen}
        }
    \end{subfigure}
    \hfill
    \begin{subfigure}{0.27\linewidth}
        \centering
        \includegraphics[width=\linewidth]{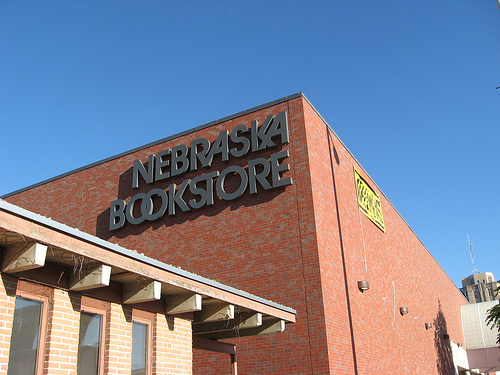}
        \caption{
        \gt{}$=$\regt{}:  \\ 
        \hphantom{(a) }\textcolor{ForestGreen}{bookstore}
        }
    \end{subfigure}
    \hfill
    \begin{subfigure}{0.205\linewidth}
        \centering
        \includegraphics[width=\linewidth]{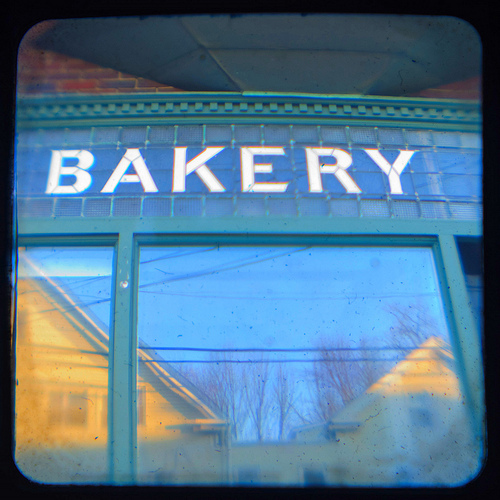}
        \caption{
        \gt{}$=$\regt{}:  \\ 
        \hphantom{(a) }\textcolor{ForestGreen}{bakery}
        }
    \end{subfigure}
    \hfill
    \begin{subfigure}{0.205\linewidth}
        \centering
        \includegraphics[width=\linewidth]{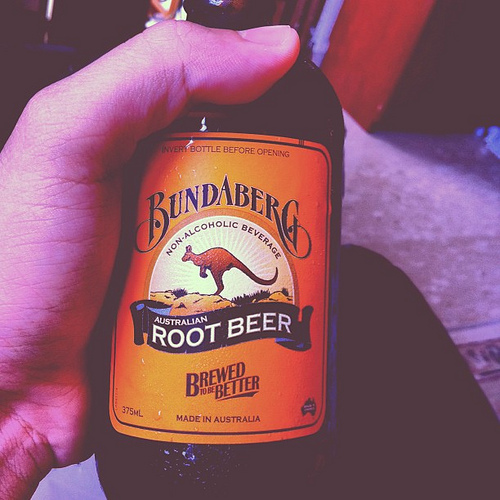}
        \caption{
        \gt{}$=$\regt{}:  \\ 
        \hphantom{(a) }\textcolor{ForestGreen}{beer bottle}
        }
    \end{subfigure}
    \caption{Images containing objects with the \attr{text-recognition} 
    attribute, where image classification relies on reading and interpreting 
    text.}
    \label{fig:ocr}
\end{figure}
\begin{figure}[t]
    \centering
    \begin{subfigure}[t]{0.31\linewidth}
        \vspace{0pt}
        \centering
        \includegraphics[width=\linewidth]{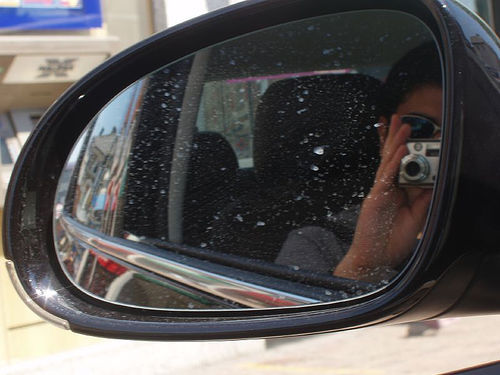}
        \caption{
        \gt{}: \textcolor{ForestGreen}{car mirror} \\ 
        \hphantom{(a) }\regt{}: \textcolor{ForestGreen}{car mirror}{; }\underline{\textcolor{ForestGreen}{sunglasses}}
        }
    \end{subfigure}
    \hfill
    \begin{subfigure}[t]{0.326\linewidth}
        \vspace{0pt}
        \centering
        \includegraphics[width=\linewidth]{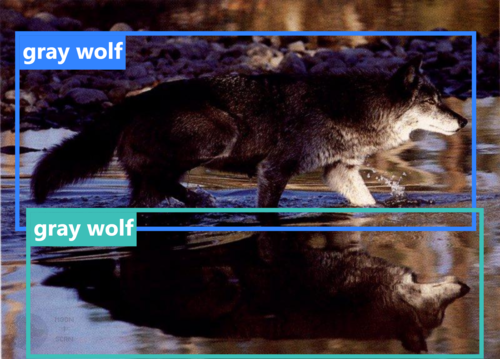}
        \caption{
        \gt{}: \textcolor{ForestGreen}{gray wolf} \\ 
        \hphantom{(a) }\regt{}: \underline{\textcolor{ForestGreen}{gray wolf}}
        }
    \end{subfigure}
    \hfill
    \begin{subfigure}[t]{0.34\linewidth}
        \vspace{0pt}
        \includegraphics[width=0.92\linewidth]{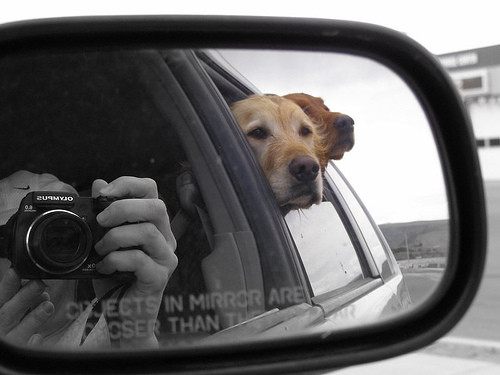}
        \caption{
        \gt{}:  \textcolor{ForestGreen}{car mirror} \\ 
        \hphantom{(d) }\regt{}: \textcolor{ForestGreen}{car mirror}{; }\underline{\textcolor{ForestGreen}{golden retriever}}
        }
    \end{subfigure}

    \caption{Images containing objects with the \underline{\attr{reflection}} attribute. This attribute is used regardless of whether the reflected object is visible (b) or not (a, c). Compact camera is not an \imnet{} class. 
    }
    \label{fig:reflected}
\end{figure}

\begin{figure}[t]
    \centering
    \begin{subfigure}[t]{0.196\textwidth}
        \vspace{0pt}
        \centering
        \includegraphics[width=\linewidth]{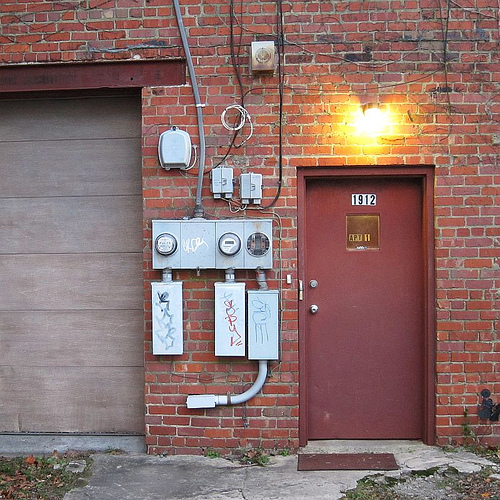}
        \caption{
        \gt{}: \textcolor{ForestGreen}{doormat} \\ 
        \hphantom{(a) }\regt{}: \textcolor{ForestGreen}{doormat}
        }
    \end{subfigure}
    \hfill
    \begin{subfigure}[t]{0.26\textwidth}
        \vspace{0pt}
        \centering
        \includegraphics[width=\linewidth]{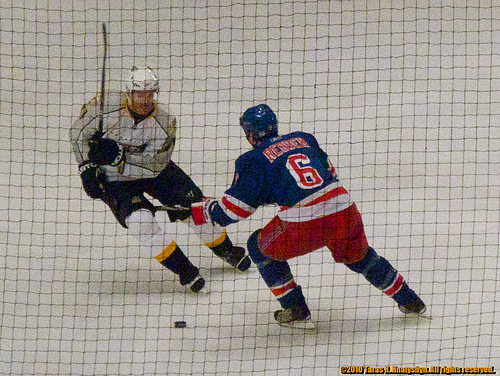}
        \caption{
        \gt{}: \textcolor{ForestGreen}{hockey puck} \\ 
        \hphantom{(a) }\regt{}: \underline{\textcolor{ForestGreen}{hockey puck}}
        }
    \end{subfigure}
    \hfill
    \begin{subfigure}[t]{0.25\textwidth}
        \vspace{0pt}
        \centering
        \includegraphics[width=\linewidth]{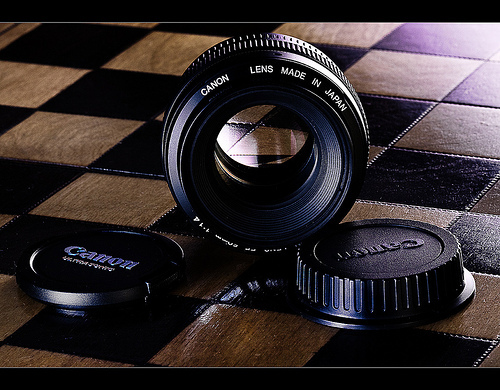}
        \caption{
        \gt{}: \textcolor{ForestGreen}{lens cap} \\ 
        \hphantom{(a) }\regt{}: \textcolor{ForestGreen}{lens cap}}
    \end{subfigure}
    \hfill
    \begin{subfigure}[t]{0.27\textwidth}
        \vspace{0pt}
        \centering
        \includegraphics[width=0.97\linewidth]{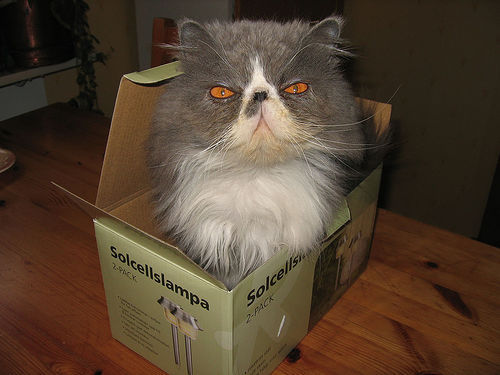}
        \caption{
        \gt{}: \textcolor{ForestGreen}{Persian cat} \\ 
        \hphantom{(a) }\regt{}: \underline{\textcolor{ForestGreen}{Persian cat}}; \textcolor{ForestGreen}{carton}
        }
    \end{subfigure}

\caption{Examples illustrating the \underline{\attr{dominant}} attribute. Dominance is entirely context-dependent and does not correlate with object size: 
(a)~\cls{doormat} occupies most of the image yet is not \attr{dominant}, 
(b)~\cls{hockey puck} is small but immediately draws attention and is marked \attr{dominant}, (c)~\cls{lens cap} is large yet not \attr{dominant} given the surrounding scene, and (d)~\cls{Persian cat} is marked \attr{dominant} despite sharing the image with \cls{carton}.}
\label{fig:saliency}
\end{figure}

\subsection{Annotator Training and Quality Control}
\label{subsec:supp-training}

\textbf{\noindent{Problematic class relationships.}}
Annotators studied problematic class relationships (identified in~\cite{kisel2024flaws}) prior to annotation. 
These include subclass--superclass overlaps, such as \cls{elephant}~$\subset$~\cls{tusker} or \cls{bathtub}~$\subset$~\cls{tub}, resolved in annotation by selecting the most specific applicable class; 
part-of relationships such as \cls{space bar}~$\subset$~\cls{keyboard}, resolved by annotating all classes present in the image; 
and near-synonym classes such as \cls{laptop}~/~\cls{notebook}, \cls{shore}~/~\cls{lakeside}, and \cls{swimsuit}~/~\cls{maillot}, resolved by selecting any one of the synonymous classes consistently (near-synonyms list builds on~\cite{beyer2020imagenet, vasudevan2022doesdoughbagelanalyzing, kisel2026multimodallargelanguagemodels}, extended through our annotation process).

\textbf{\noindent{Attribute and multi-label object examples.}}
Annotators worked through approved annotated examples for each \attr{attribute}, similar to those shown in~\Cref{fig:rendition,fig:crowd,fig:ocr,fig:reflected,fig:saliency}. 
Besides, they were familiarised with the existence of multi-label objects (\ie single objects belonging to multiple non-hierarchically related classes, therefore assigned with multiple labels). 
Classes that frequently require multi-label annotation include materials like \cls{wool} and \cls{velvet}, more examples are shown in \Cref{fig:multi}.

\textbf{\noindent{Training classes and quality assessment.}}
Annotators gained hands-on experience with the task and familiarised themselves with the annotation app using a set of intentionally challenging training classes chosen by the authors: visually similar concepts that are easily confused (\eg not every camera is a \cls{reflex camera}), uncommon concepts that also tested localisation decisions (\eg how to draw a bounding box for \cls{spotlight}), and images in which multiple ImageNet classes appear simultaneously. 
The authors manually reviewed these training annotations against a predefined quality threshold. 
Annotators who met it proceeded to real tasks; those who did not received targeted feedback and additional training classes, after which the review process was repeated.

\textbf{\noindent{Control sets.}}
To monitor annotation consistency beyond the initial training phase, we periodically 
assigned the same image sets to multiple annotators (who did not know it was a control set, but sometimes found out when discussing problems) and compared their outputs. 
Control sets typically consisted 
of 2--3 carefully selected classes, often from the same class group 
(see~\Cref{subsec:workflow} and Suppl.~\Cref{subsec:supp-groups}). 
Annotators who fell below the expected standard received targeted feedback and were asked to revisit their annotations.
Beyond individual control, this process revealed edge cases the guidelines had not yet covered, playing an important role in the iterative protocol updates described in~\Cref{subsec:iterative}.

\subsection{Class Group Details}
\label{subsec:supp-groups}

Classes are organised into 97 groups ranging from 1 to 59 classes, with a median of 7 and an average of 10 classes per group. 
Groups were constructed by clustering classes that share a common parent node in the WordNet hierarchy. 
However, they required substantial manual revision: WordNet parent assignments frequently produced semantically incoherent groupings, and the granularity of the hierarchy is highly unbalanced. 
This is unsurprising given that \imnet{} classes were scraped as WordNet leaf nodes, but they still share hierarchical relationships (one class being a subset of another, or classes being near-synonyms~\cite{beyer2020imagenet,vasudevan2022doesdoughbagelanalyzing,kisel2024flaws}).

The largest group, \cls{birds} (59 classes), is an outlier: unlike dogs, which are naturally split into fine-grained subgroups (\eg terriers, hounds, retrievers), birds lack a similarly clean taxonomic subdivision at the \imnet{} level (or the authors lack the domain knowledge to identify one).

\subsection{Per-Class Preparation Protocol}
\label{subsec:per-class-supp}

We describe the per-class preparation protocol each annotator was instructed to follow before annotating any image in their assigned group. Annotators were not time-constrained; it was up to their judgment to decide when they were familiar enough with the subject to start annotating.

\begin{figure}[t]
    \centering
    \centering
    \begin{subfigure}{0.23\linewidth}
        \includegraphics[width=0.8\linewidth]{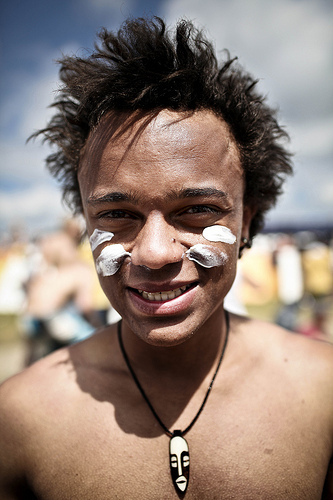}
        \caption{
        \gt{}: \textcolor{red}{sunscreen} \\ 
        \hphantom{(a) }\regt{}: no valid label
        }
    \end{subfigure}
    \hspace{2em}
    \begin{subfigure}{0.26\linewidth}
        \includegraphics[width=0.715\linewidth]{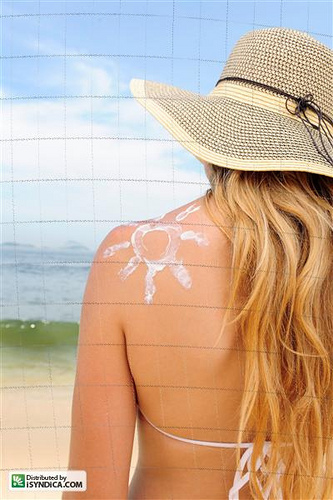}
        \caption{
        \gt{}: \textcolor{ForestGreen}{sunscreen} \\ 
        \hphantom{(a) }\regt{}: \textcolor{ForestGreen}{sunscreen}{; } 
        \textcolor{ForestGreen}{bikini}
        }
    \end{subfigure}
    \caption{
    The role of context in image recognition. Both images show a cream-like white substance on skin. In~(a), the festival-like background and characteristic stripes on the face suggest it is body paint. In~(b), a woman in a bikini on the beach and wearing a sun hat suggests it is \cls{sunscreen}. Either interpretation could apply to both images. We can only make an educated guess based on the context.}
    \label{fig:context}
\end{figure}

\textbf{\noindent{Examining class content.}}
Annotators first examined the actual image content of each class in their assigned group rather than relying on class names alone --- the inverse of the original \imnet{} procedure~\cite{deng2009imagenet}, where crowd workers were 
shown an image and asked to verify a given label. 
This step was necessary because ImageNet class names frequently misrepresent their content: the original annotation process often diverged a class entirely from its WordNet definition (\eg \cls{tiger cat} in practice contains striped tabby cats, oncillas, and tigers~\cite{kisel2024flaws}).

The preparation involved consulting external references (\eg Wikipedia, iNaturalist) to create reliable visual identification criteria for each class. 
Annotators identified the key distinguishing features that help with recognition (\eg for a snake class, features such as head shape, body pattern, or scale texture), detected common confusions (\eg two insect classes that belong to the same order but differ in antenna length and body shape), and verified that the actual images in the class match the expected visual content. 

\textbf{\noindent{Class definition writing.}}
Based on the previous examination, the annotator recorded a working definition for each class in a shared table, grounded in actual image content rather than the WordNet label (the \imnet{} basis for class definitions, which was often lost 
during data collection and labelling). 
For unambiguous classes (\eg \cls{tennis ball}) this was brief; for problematic ones this could take considerably longer. 
For instance, \cls{ear, spike, capitulum} nominally refers to grain spikes by its WordNet definition, but the overwhelming majority of images depict \cls{corn} (in the American sense) in various forms --- an obvious discrepancy that must be resolved before annotation begins.

When the class name was misleading or the image content was highly diverse, the definition was settled collectively in a supervised group chat in which the authors actively participated. 
Solutions were individual to each class: for unambiguous mismatches we adopted a renaming; for classes with mixed or ambiguous content the resolution was more involved. 
Throughout this process, we maintained a strong preference for preserving original images over discarding them --- we would rather revise a class name to better reflect the actual image content than remove images from the dataset. 
Our per-class definitions are publicly available and will be updated as further issues are identified by the community (see~\Cref{subsec:limitations}).
    
    

\begin{figure}[t]
    \centering
    \includegraphics[width=\linewidth]{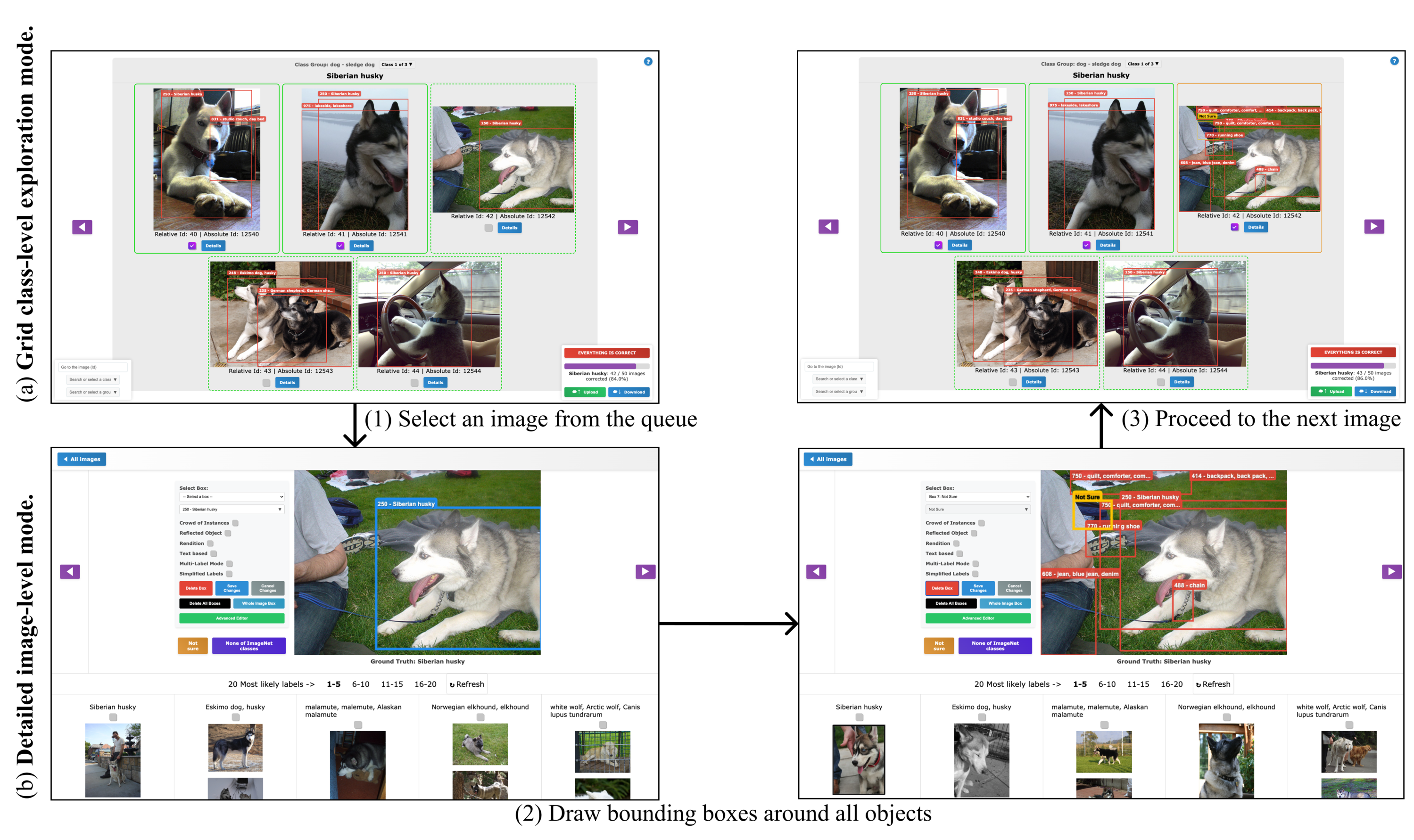}
    \caption{Annotation application workflow. Screenshots illustrate 
    the annotation loop supported by the application. (1)~Select an 
    image from the task queue, organized by class and browsable in a grid view 
    for class-level inspection and definition review. (2)~Switch to 
    the image-level mode to draw bounding boxes, add \attr{attributes}, edit existing annotations, and reference model predictions. (3)~After completing 
    the current image, proceed to the next, repeating the loop.}
    \label{fig:pipeline}
\end{figure}
\begin{figure}[t]
    \centering
    \begin{subfigure}{0.49\linewidth}
        \centering
        \includegraphics[width=\linewidth]{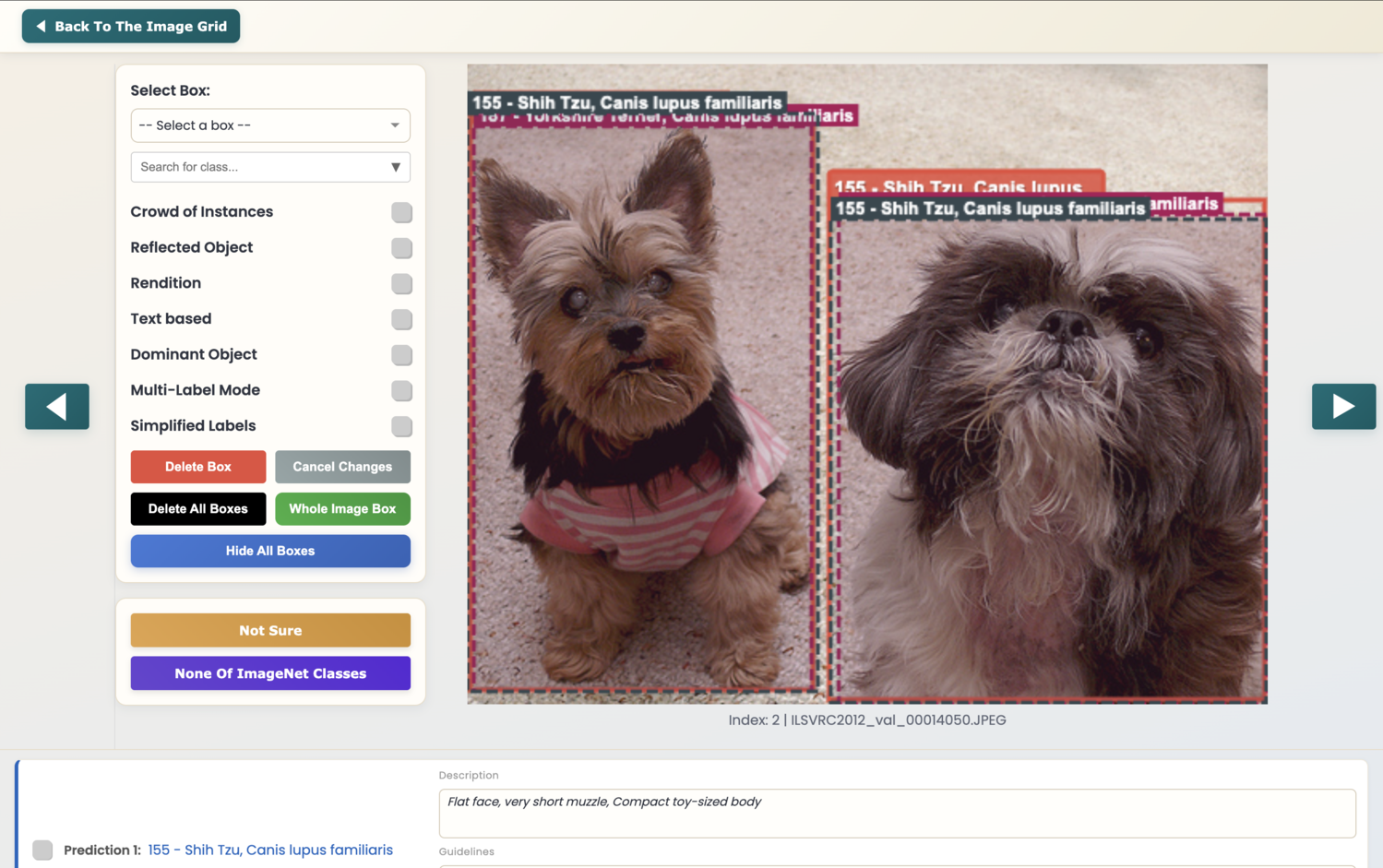}
        \caption{Image-level view with \ac{mllm}+SAM3 proposals shown as 
        dashed bounding boxes.}
    \end{subfigure}
    \hfill
    \begin{subfigure}{0.49\linewidth}
        \centering
        \includegraphics[width=\linewidth]{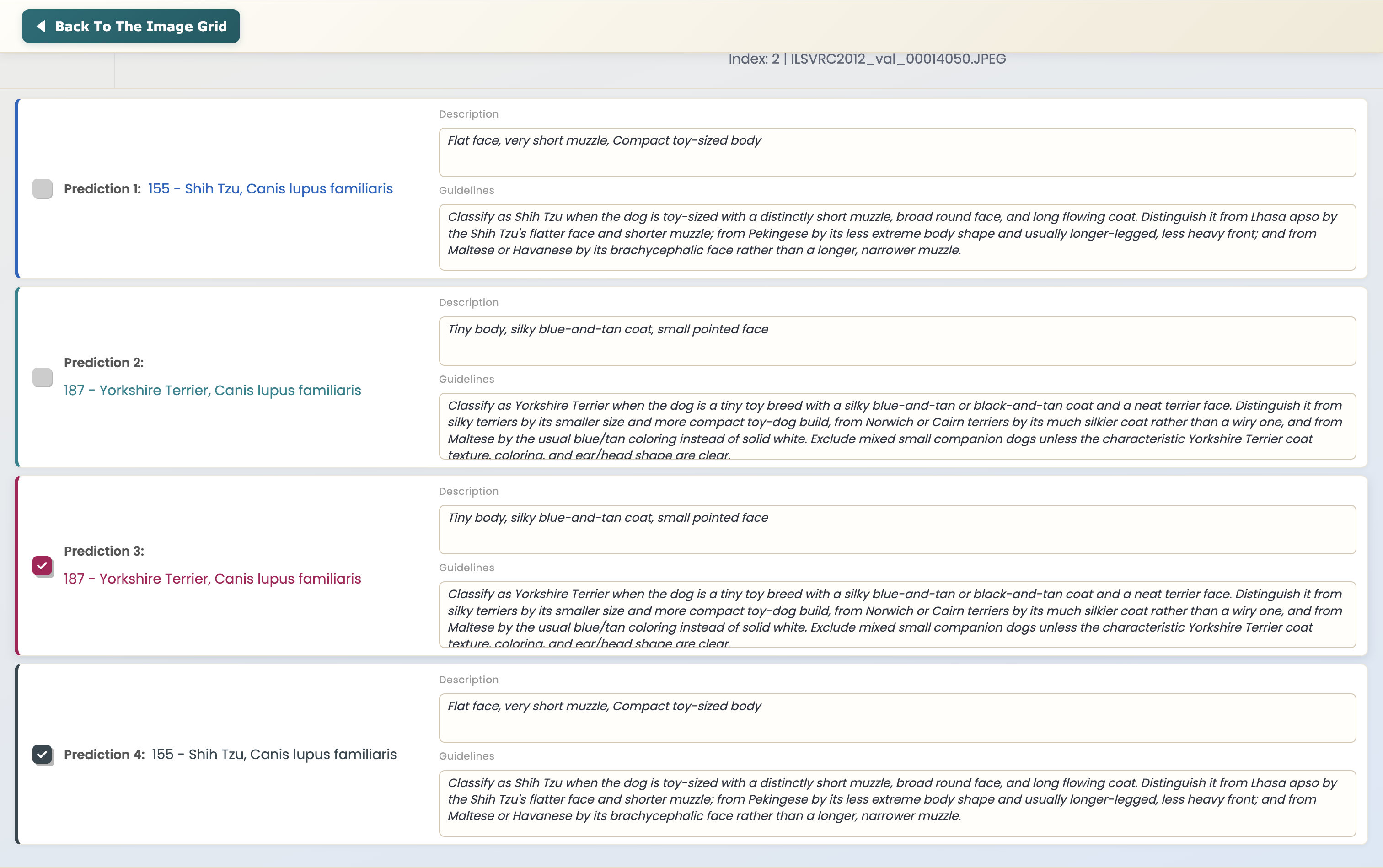}
        \caption{Prediction panel with class descriptions and annotation 
        guidelines for each proposed class.}
    \end{subfigure}
    \caption{Annotation application update for the verification phase. 
    The core functionality remains the same as in the main phase; the key 
    changes are the replacement of OWLv2/OpenCLIP predictions with 
    \ac{mllm}+SAM3 proposals and the addition of class definitions 
    directly in the prediction panel.}
    \label{fig:app-verification}
\end{figure}

\subsection{Annotation App}
\label{subsec:supp-app}

We developed a custom web-based annotation application designed to support the pipeline described in~\Cref{subsec:workflow}. 
The application integrates model predictions (OWLv2 bounding box proposals and OpenCLIP image predictions in the main phase; 
\ac{mllm} predictions 
with SAM3 localisations in the verification phase), per-class definitions, and cross-class navigation into a single interface designed to support both class-level review and per-image 
annotation. 
It provides two complementary views: a grid-based 
class-level view and a detailed image-level view.

\noindent{\textbf{Main phase.}} Work begins in grid mode, where annotators browse multiple images per class and propose class definitions as described in~\Cref{subsec:workflow,subsec:per-class-supp}. 
The grid displays images in batches, allowing annotators to quickly assess the overall content of a class and check existing annotations in context. 
A progress bar indicates how many images in the current class have been annotated. Annotators can navigate between classes within their assigned group via a dropdown menu. 
Each image in the grid shows a checkmark indicating whether it has been annotated.

After inspecting the classes, annotators proceed to the image-level labelling workflow, illustrated in~\Cref{fig:pipeline}. 
Bounding boxes are drawn by holding the left mouse button and can be selected, resized by dragging their edges, or deleted. 
Each selected box can be assigned a class label via a dropdown menu, and one or more \attr{attributes} can be toggled via checkboxes or keyboard shortcuts.
When an object belongs to multiple non-hierarchically related classes (see~\Cref{fig:multi}), annotators use multi-label mode to assign several labels to a single box rather than 
drawing separate boxes.

If an annotator is uncertain about the class of an object, they can mark it using the \emph{Not sure} button, optionally selecting one or more candidate classes. 
Such boxes are highlighted in yellow and revisited during the second verification phase (see~\Cref{subsec:iterative}). 
If no \imnet{} class is present in the image at all, annotators select \emph{None of ImageNet classes} instead. 
The \emph{Not sure} option is preferred over this when any doubt remains, to avoid missing valid objects and discarding images from the dataset.

Below the image editor, the top-20 OpenCLIP class predictions are displayed alongside example images from the dataset for each predicted class. 
Annotators can zoom into any example image for a closer look and use a refresh button to load alternative examples. 
If a predicted class appears to be present in the image, annotators can check its checkbox so that the next drawn box is automatically assigned to that class. 
Annotations are saved automatically whenever the annotator navigates to another image or returns to the grid. 
Annotators were instructed to upload their work to a shared storage at the end of each working session using the \emph{Upload} button in grid mode.

\noindent{\textbf{In the verification phase}} (see~\Cref{subsec:iterative}), the annotation application was updated to 
support the new workflow while retaining the same core 
functionality. 
The main changes are illustrated in~\Cref{fig:app-verification}.

OWLv2 bounding box proposals and OpenCLIP top-20 predictions from the main phase are replaced by \ac{mllm} predictions with SAM3-generated localisations~\cite{carion2026sam3segmentconcepts}. 
These appear as dashed bounding boxes overlaid on the image (\Cref{fig:app-verification}, left); 
annotators can click on a dashed box to accept and copy it if they agree with the model proposal. 
Below the image, each prediction is displayed alongside the corresponding class description and annotation guidelines (\Cref{fig:app-verification}, right), allowing annotators to verify predictions against the shared class definitions directly within the interface. 
Annotators can check the prediction checkbox to reveal the localisation for the proposed box (the previously mentioned dashed boxes).

\begin{figure}[p]
    \centering

    \begin{subfigure}{\linewidth}
        \centering
        \includegraphics[width=0.8\linewidth]{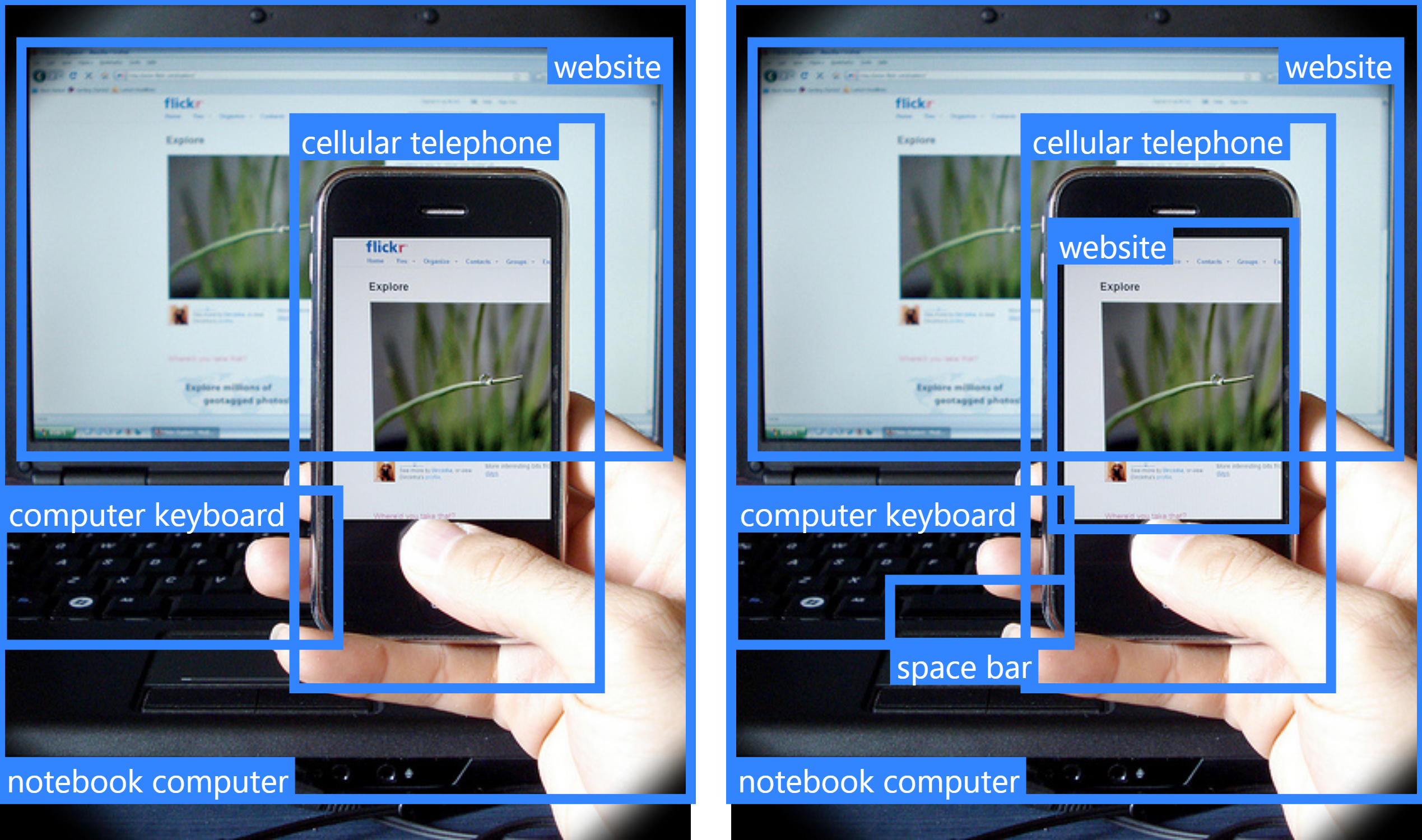}
    \end{subfigure}

    \vspace{0.5em}
    
    \begin{subfigure}{\linewidth}
        \centering
        \includegraphics[width=0.8\linewidth]{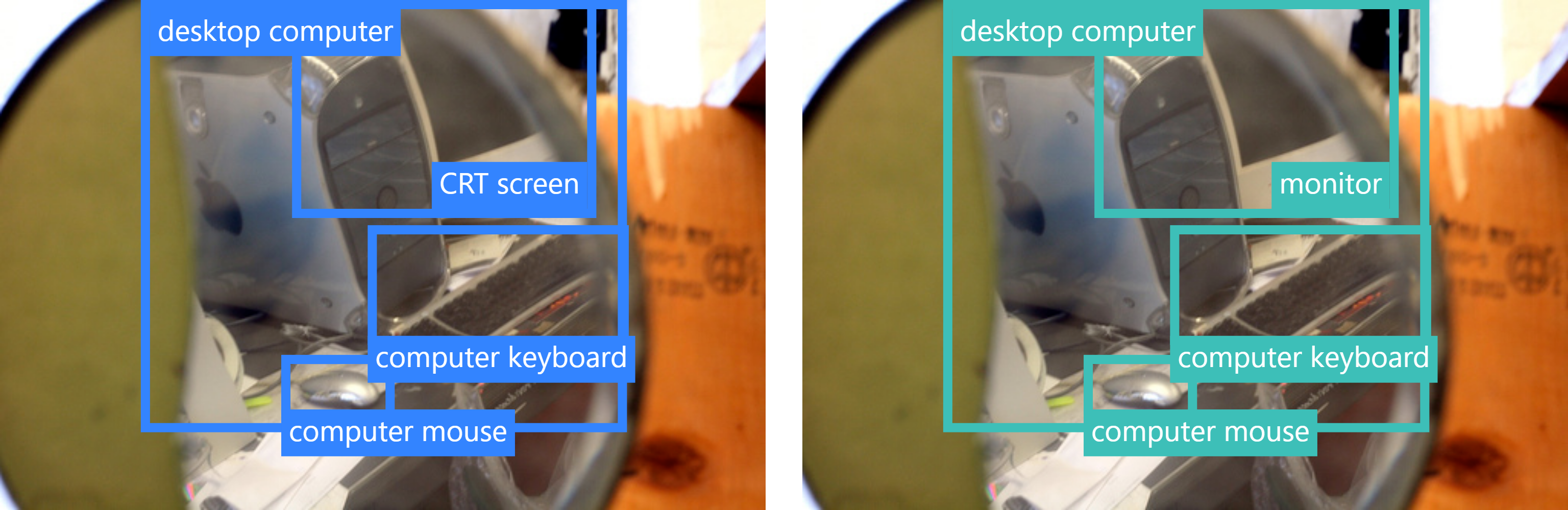}
    \end{subfigure}
    
    
    \vspace{0.5em}
    
     \begin{subfigure}{\linewidth}
        \centering
        \includegraphics[width=0.8\linewidth]{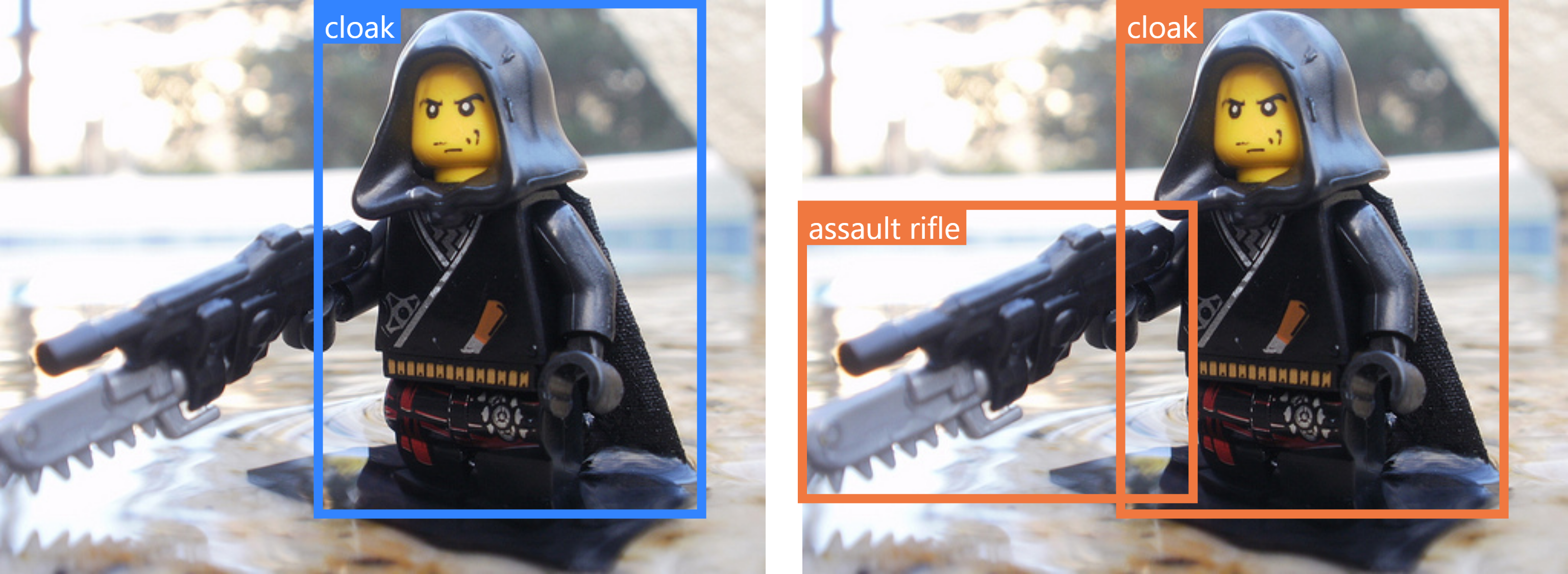}
    \end{subfigure}
    
    \vspace{0.5em}
    
    \begin{subfigure}{\linewidth}
        \centering
        \includegraphics[width=0.8\linewidth]{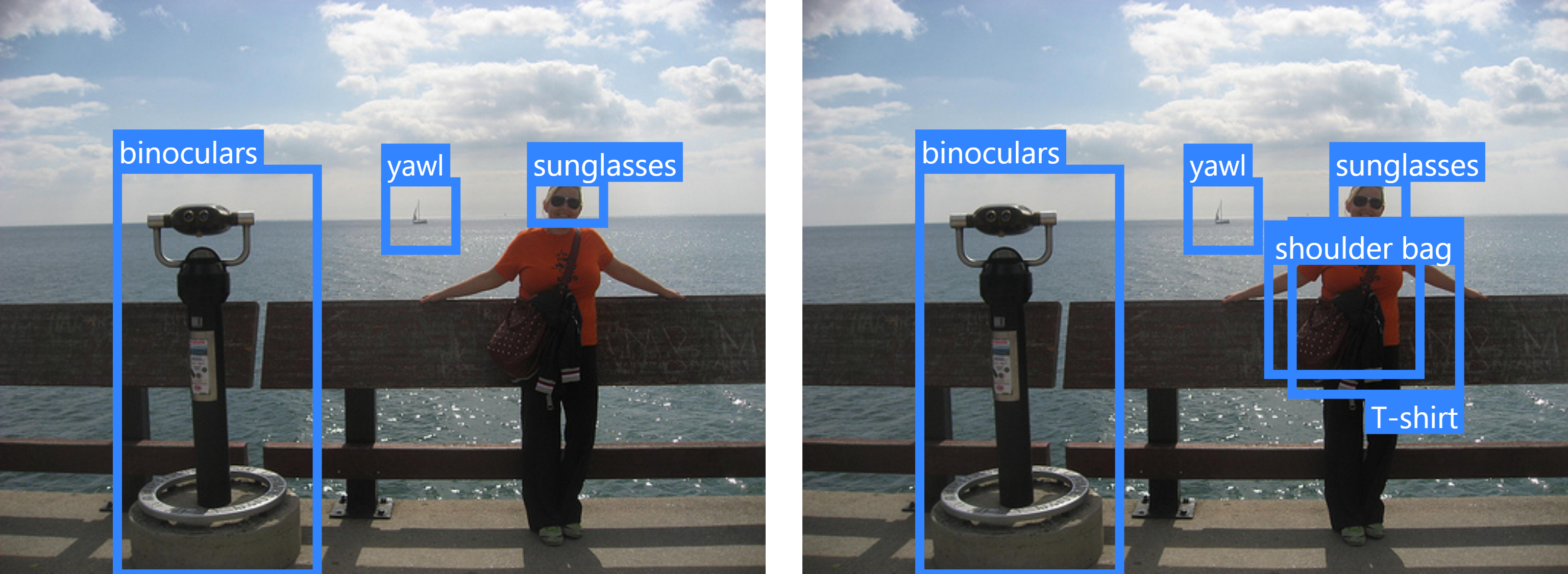}
    \end{subfigure}
    
    \caption{Examples of annotation changes made during the second 
    verification phase. Each row shows a before (left) and after (right) 
    pair. Colors denote objects with:
    \textcolor[HTML]{3384FF}{\rule{1.2ex}{1.2ex}}~no attributes,
    \textcolor[HTML]{F07840}{\rule{1.2ex}{1.2ex}}~\attr{rendition} attribute, 
    \textcolor[HTML]{3DBFB8}{\rule{1.2ex}{1.2ex}}~\attr{reflection} attribute. 
    Annotators add previously unnoticed objects, verify that existing annotations correspond to finalised class definitions, and assign attributes that were introduced or clarified mid-annotation.}
    \label{fig:verification}
\end{figure}


\clearpage

\section{Experimental setup details}
\label{sec:supp_setup}

\textbf{\noindent{Model pool.}}
The full list of models used in the paper for evaluation, with their reference naming to the exact model version. \cgpt{} is the model used through the OpenAI Responses API. \qwen{} refers to the \texttt{Qwen3-VL-235B-A22B-it model}. \third{} is the \texttt{Gemma-4-31B-it}. \siglip{}, \sigliptwo{}, \sigliptwog{} refer to \texttt{SigLIP so400M/14-384}, \texttt{SigLIP~2 so400M/16-384}, and \texttt{SigLIP~2 ViT/g-opt-384} versions. \dino{} refers to the \texttt{DINOv3 ViT-7B} encoder used to obtain image embeddings and perform $k$-NN search in the embedding space; the best $k$=11 is selected for all experiments based on \regt{} accuracy. \effnetv{}, \effnetl{} are supervised models referring to \texttt{EfficientNetV2-XL} and \texttt{EfficientNet-L2}. The best Timm \cite{rw2019timm} leaderboard \imnet{} model \texttt{EVA-02 ViT-L-14-448} is used and referred to as EVA-02.

\textbf{\noindent{\acp{mllm} image classification.}}
Leveraging \acp{mllm} image classification knowledge, the setup commonly referred to as ``OW'' in the field is selected for all experiments. A single prompt is used for all \acp{mllm}, matching exactly the one located in the supplementary of our prior paper~\cite{kisel2026multimodallargelanguagemodels}. The encoder choice matches exactly our prior work too, utilizing the best-performing encoder per model \texttt{Qwen3-Embedding-8B / SigLIP~2 ViT-gopt-384 (text encoder)}. A class name templating trick following~\cite{radford2021learningtransferablevisualmodels} is used before tokenization for both \acp{mllm} and \acp{vlm}.

\textbf{\noindent{Computational resources.}}
All experiments were conducted on the local university server. The required computational resources varied considerably across tasks: smaller experiments utilized a single NVIDIA A100 SXM4 40 GB GPU on an AMD EPYC 7543/7763 CPU cluster, while the largest model in our experiments: \texttt{Qwen3-VL-235B-A22B} required 4-6 NVIDIA HGX H200 GPUs on an AMD EPYC 9355 CPU cluster. Experiment runtimes ranged from approximately 10 minutes to a maximum of 72 hours.

\clearpage

\section{Additional Results}
\label{sec:supp_additional_results}

\begin{table*}[ht]
\footnotesize
\centering
\caption{Top-1 ReGT accuracy on $M$ by ground-truth bounding box area. Index $i$ denotes the $i$-th largest box. Columns $i_n$ report accuracy for the $i$-th largest box over $n$ images. Columns $\mathrel{\scriptstyle \le}i_n$ report ReaL accuracy, treating the top-$i$ boxes as valid labels. $\lim_{i \to \infty} \mathrm{Acc}_{\le i}(subset) = \mathrm{Acc}_{\mathrm{ReGT}}(M)$. }
\setlength{\tabcolsep}{2.5pt}
\begin{tabular}{l cccccc c ccccc}
\toprule

& \multicolumn{6}{c}{$i$-th largest bounding box} & & \multicolumn{5}{c}{top~$i$ largest bounding boxes} \\
\cmidrule{2-7} \cmidrule{9-13}
Model & $Dom_{668}$ &
$1_{\cntReGTM}$ & 
$2_{15754}$ & 
$3_{9343}$ & 
$4_{5867}$ & 
$5_{3840}$ & &
$\mathrel{\scriptstyle \le}\!2_{\cntReGTM}$ & 
$\mathrel{\scriptstyle \le}\!3_{\cntReGTM}$ & 
$\mathrel{\scriptstyle \le}\!4_{\cntReGTM}$ & 
$\mathrel{\scriptstyle \le}\!5_{\cntReGTM}$ & 
$\mathrm{M}_{\cntReGTM}$ \\

\midrule
\cgpt{} & 78.14 & 51.23 & 26.51 & 16.80 & 11.22 & 7.71 & & 72.13 & 75.83 & 76.69 & 77.00 & 77.22 \\
\qwen{} & 75.60 & 52.78 & 26.62 & 16.63 & 11.04 & 8.02 & & 73.67 & 77.47 & 78.34 & 78.70 & 79.08 \\
\third{} & 79.34 & 49.50 & 26.27 & 16.76 & 11.42 & 7.97 & & 70.18 & 74.12 & 74.99 & 75.32 & 75.60 \\

\midrule

\siglip{} & 76.35 & 57.04 & 30.48 & 18.78 & 12.63 & 8.65 & & 81.10 & 85.11 & 85.95 & 86.28 & 86.60 \\
\sigliptwo{} & 78.74 & 57.28 & 30.84 & 18.68 & 12.43 & 8.98 & & 81.60 & 85.58 & 86.46 & 86.82  & 87.16 \\
\sigliptwog{} & 78.59 & 57.46 & 30.99 & 18.84 & 12.61 & 8.65 & & 81.95 & 86.09 & 86.98 & 87.30 & 87.66 \\

\midrule

\dino{} & 72.60 & 55.83 & 30.27 & 18.57 & 12.61 & 9.14 & & 79.62 & 83.63 & 84.55 & 84.96 & 85.28 \\
\effnetv{} & 70.81 & 56.61 & 31.30 & 19.07 & 12.54 & 9.32 & & 81.38 & 85.53 & 86.45 & 86.85 & 87.18 \\
\effnetl{} & 76.20 & 58.30 & 32.15 & 20.08 & 13.74 & 9.87 & & 83.72 & 88.17 & 89.22 & 89.68 & 90.08 \\
\eva{} & 76.65 & 58.80 & 32.72 & 19.97 & 13.65 & 10.03 & & 84.68 & 89.17 & 90.21 & 90.70 & 91.11 \\

\bottomrule
\end{tabular}
\label{tab:merged_bbox_stats_reordered}
\end{table*}

\textbf{Bounding box area analysis} is shown in~\Cref{tab:merged_bbox_stats_reordered}.
For each multilabel image, we evaluate model predictions against the label of 
the $i$-th largest bounding box with a unique label, converting multilabel 
images into effectively single-label ones. The subset size for $i{=}1$ equals 
$\cntReGTM$ (all $\mathrm{M}$ images); it drops to $15{,}754$ for $i{=}2$ 
because some images contain a single box annotated with multiple labels (see~\Cref{fig:size}), 
leaving no second uniquely-labelled box. For higher $i$ the subsets shrink 
further simply because fewer images contain that many boxes. Accuracy falls 
sharply with box size -- all models score 50--59\% against the largest box, 
but only 8--10\% for the fifth -- confirming a strong saliency bias. 
However, the cumulative columns show that models do not ignore smaller objects 
entirely: the largest gain appears between 
$\mathrel{\scriptstyle \le}\!1$ and $\mathrel{\scriptstyle \le}\!2$, with 
continued but diminishing returns for deeper ranks, converging toward full 
$\mathrm{M}$ accuracy. The $Dom$ column evaluates against the 
union of labels on \attr{dominant}-flagged boxes per image. \acp{mllm} and 
\acp{vlm} score consistently around 76--79\%, while supervised models vary 
more (EffNetV2: 71\%, EVA-02: 77\%), suggesting \acp{mllm} and \acp{vlm} are 
better calibrated to human saliency judgments.

\clearpage

\begin{table*}[t]
\footnotesize
\centering
\begin{minipage}[t]{0.52\linewidth}
    \caption{Top-1 accuracy by number of unique \regt{} labels per image. Each column $k_n$ denotes a subset of $k$-label images of size $n$, with the last column grouping images with $\mathrel{\scriptstyle \ge}\!6$ labels.}
    \setlength{\tabcolsep}{2.5pt}
    \begin{tabular}{lc@{\hskip 0.3cm}rrrrr}
    \toprule
    & $\mathrm{S}_{\cntReGTS}$ & \multicolumn{5}{c}{$\mathrm{M}_{\cntReGTM}$}\\ 
    \cmidrule(r){2-2} \cmidrule{3-7} 
    Model & $\mathrm{1}_{\cntReGTS}$ & $\mathrm{2}_{8463}$ & $\mathrm{3}_{4082}$ & $\mathrm{4}_{2103}$ & $\mathrm{5}_{1048}$ & $\mathrel{\scriptstyle \ge}\!6_{970}$ \\
    \midrule
    \cgpt{} & \cellcolor{acccolor!36}77.86 & \cellcolor{acccolor!33}76.69 & \cellcolor{acccolor!33}76.87 & \cellcolor{acccolor!39}78.60 & \cellcolor{acccolor!36}78.53 & \cellcolor{acccolor!35}78.97 \\
    \qwen{} & \cellcolor{acccolor!32}75.97 & \cellcolor{acccolor!36}78.52 & \cellcolor{acccolor!35}78.78 & \cellcolor{acccolor!41}80.88 & \cellcolor{acccolor!36}78.44 & \cellcolor{acccolor!43}81.96 \\
    \third{} & \cellcolor{acccolor!22}74.48 & \cellcolor{acccolor!24}75.23 & \cellcolor{acccolor!24}75.06 & \cellcolor{acccolor!32}78.03 & \cellcolor{acccolor!23}75.00 & \cellcolor{acccolor!28}76.49 \\
    \midrule
    \siglip{} & \cellcolor{acccolor!49}88.54 & \cellcolor{acccolor!42}86.00 & \cellcolor{acccolor!45}86.84 & \cellcolor{acccolor!47}88.02 & \cellcolor{acccolor!47}87.40 & \cellcolor{acccolor!44}86.80 \\
    \sigliptwo{} & \cellcolor{acccolor!50}89.11 & \cellcolor{acccolor!45}87.11 & \cellcolor{acccolor!44}86.80 & \cellcolor{acccolor!49}88.83 & \cellcolor{acccolor!43}86.35 & \cellcolor{acccolor!44}86.39 \\
    \sigliptwog{} & \cellcolor{acccolor!51}89.69 & \cellcolor{acccolor!47}87.65 & \cellcolor{acccolor!46}87.58 & \cellcolor{acccolor!48}88.21 & \cellcolor{acccolor!45}87.02 & \cellcolor{acccolor!47}87.63 \\
    \midrule
    \dino{} & \cellcolor{acccolor!49}88.97 & \cellcolor{acccolor!43}86.27 & \cellcolor{acccolor!39}84.76 & \cellcolor{acccolor!38}84.07 & \cellcolor{acccolor!33}82.54 & \cellcolor{acccolor!38}84.43 \\
    \effnetv{} & \cellcolor{acccolor!49}88.78 & \cellcolor{acccolor!45}86.99 & \cellcolor{acccolor!45}87.16 & \cellcolor{acccolor!48}88.21 & \cellcolor{acccolor!45}87.12 & \cellcolor{acccolor!45}86.80 \\
    \effnetl{} & \cellcolor{acccolor!55}90.76 & \cellcolor{acccolor!53}89.77 & \cellcolor{acccolor!54}90.45 & \cellcolor{acccolor!56}91.16 & \cellcolor{acccolor!50}89.03 & \cellcolor{acccolor!53}90.10 \\
    \eva{} & \cellcolor{acccolor!57}91.52 & \cellcolor{acccolor!54}90.72 & \cellcolor{acccolor!55}91.03 & \cellcolor{acccolor!59}92.34 & \cellcolor{acccolor!55}90.94 & \cellcolor{acccolor!69}92.27 \\
    \bottomrule
    \end{tabular}
    \label{tab:multilabel_stats}
\end{minipage}
\hfill
\begin{minipage}[t]{0.44\linewidth}
    \caption{Single-label compatible evaluation protocols defined in the \Cref{sec:results} and Suppl.~\Cref{sec:supp_additional_results}. Subscripts denote the number of images available for each protocol.}
    \setlength{\tabcolsep}{2.5pt}
    \begin{tabular}{l ccc}
    \toprule
    & \multicolumn{3}{c}{\regt{}} \\
    \cmidrule{2-4}
    Model & $\mathrm{S}_{\cntReGTS}$ & $Dom'_{323}$ & $ExCrops_{38628}$ \\
    \midrule
    \cgpt{} & 77.86 & 79.57 & - \\
    \qwen{}  & 75.97 & 67.80 & - \\
    \third{} & 74.48 & 82.04 & 65.78 \\
    \midrule
    \siglip{} & 88.54 & 72.45 & 77.34 \\
    \sigliptwo{} & 89.11 & 76.47 & 77.95 \\
    \sigliptwog{} & 89.69 & 75.85 & 78.87 \\
    \midrule
    \dino{} & 88.97 & 71.52 & 74.41 \\
    \effnetv{} & 88.78 & 63.16 & 76.67 \\
    \effnetl{} & 90.76 & 73.37 & 77.28 \\
    \eva{} & 91.52 & 74.30 & 77.81 \\
    \bottomrule
    \end{tabular}
    \label{tab:s_versus_dom}
\end{minipage}
\end{table*}

\noindent\textbf{Label count analysis} is presented in~\Cref{tab:multilabel_stats}. 
Accuracy remains broadly stable as the number of \regt{} labels per 
image increases, contrary to the expectation that more labels would 
create a more target-rich environment. Images with high label counts 
tend to depict complex scenes with many small objects, which offsets 
any advantage from having more valid prediction targets.

\noindent\textbf{Single-label compatible protocols} are summarized in~\Cref{tab:s_versus_dom}. 
We provide three evaluation protocols compatible with classical 
single-label ImageNet evaluation. $S$ is the standard \regt{} 
single-label accuracy baseline. $Dom'$ evaluates on multilabel images where all \attr{dominant} objects 
share the same label, counting a prediction correct if it matches that 
label. This is a stricter setting than in~\Cref{tab:merged_bbox_stats_reordered}, 
where a prediction is correct if it matches the label of any object with the \attr{dominant} attribute. $ExCrops$ evaluates on non-overlapping expanded crops, 
each guaranteed to contain exactly one annotated object 
(see~\Cref{subsec:crops}). $ExCrops$ results for \cgpt{} and \qwen{} have been omitted due to computational resource constraints.

\clearpage

\section{Comparison with Prior Reannotations}
\label{sec:supp_prior_work}

\begin{table*}[ht]
\small
\centering
\caption{Model accuracy evaluated against four prior reannotation datasets, 
each shown with original labels (top row per model) and with \regt{} label 
categories applied (bottom row). Subsets are defined relative to each dataset own annotations. Images with no valid label are excluded from evaluation, according to the label set of each row (top: images unlabeled in the prior dataset; bottom: $\mathrm{N}$ category images in \regt{}).}
\setlength{\tabcolsep}{4pt}
\begin{tabular}{lrlrrrrrrr}
\toprule
& \multicolumn{1}{c}{ImGT} & & \multicolumn{7}{c}{ImageNet ReaL~\cite{beyer2020imagenet}} \\ \cmidrule{2-2} \cmidrule{4-10} 
& \multicolumn{1}{c}{$\mathrm{A_{50000}}$} & & \multicolumn{1}{c}{$\mathrm{{A{\setminus}N}_{46837}}$} & \multicolumn{1}{c}{$\mathrm{S_{39394}}$} & \multicolumn{1}{c}{$\mathrm{{S+}_{35811}}$} & \multicolumn{1}{c}{$\mathrm{{S-}_{3583}}$} & \multicolumn{1}{c}{$\mathrm{M_{7443}}$} & \multicolumn{1}{c}{$\mathrm{{M+}_{6495}}$} & \multicolumn{1}{c}{$\mathrm{{M-}_{948}}$} \\
\cmidrule{1-10}

\cgpt{} & 71.32 & & 77.53 & 78.83 & 82.96 & 37.54 & 70.68 & 74.67 & 43.35 \\
\cgpt{} \regt{} & 71.32 & & 77.64 & 81.91 & 84.10 & 58.39 & 65.23 & 67.11 & 52.15 \\

\qwen{} & 69.77 & & 76.11 & 76.43 & 80.49 & 35.86 & 74.43 & 78.60 & 45.89 \\
\qwen{} \regt{} & 69.77 & & 77.04 & 80.38 & 82.44 & 58.27 & 69.18 & 71.05 & 56.11 \\

\third{} & 68.33 & & 74.69 & 75.65 & 79.41 & 38.12 & 69.57 & 73.49 & 42.72 \\
\third{} \regt{} & 68.33 & & 74.87 & 78.76 & 80.84 & 56.46 & 64.24 & 66.04 & 51.71 \\

\eva{} & 90.73 & & 91.39 & 92.02 & 98.18 & 30.45 & 88.08 & 95.55 & 36.92 \\ 
\eva{} \regt{} & 90.73 & & 91.38 & 93.94 & 96.03 & 71.49 & 83.37 & 85.87 & 65.90 \\ 

\toprule
& \multicolumn{1}{c}{ImGT} & & \multicolumn{7}{c}{Multi-label annotations~\cite{pmlr-v119-shankar20c}} \\ \cmidrule{2-2} \cmidrule{4-10} 
& \multicolumn{1}{c}{$\mathrm{A_{20000}}$} & & \multicolumn{1}{c}{$\mathrm{{A{\setminus}N}_{19549}}$} & \multicolumn{1}{c}{$\mathrm{S_{15892}}$} & \multicolumn{1}{c}{$\mathrm{{S+}_{15504}}$} & \multicolumn{1}{c}{$\mathrm{{S-}_{388}}$} & \multicolumn{1}{c}{$\mathrm{M_{3657}}$} & \multicolumn{1}{c}{$\mathrm{{M+}_{3588}}$} & \multicolumn{1}{c}{$\mathrm{{M-}_{69}}$} \\
\cmidrule{1-10}

\cgpt{} & 71.20 & & 78.20 & 78.44 & 78.81 & 63.66 & 77.19 & 77.42 & 65.22 \\
\cgpt{} \regt{} &  71.20 & & 77.39 & 79.96 & 80.38 & 62.12 & 69.56 & 69.73 & 60.61 \\

\qwen{} & 69.39 & & 76.88 & 76.39 & 76.78 & 60.82 & 79.00 & 79.21 & 68.12 \\
\qwen{} \regt{} & 69.39 & & 76.65 & 78.67 & 79.02 & 64.07 & 71.28 & 71.40 & 65.15 \\

\third{} & 68.09 & & 75.23 & 75.12 & 75.44 & 62.37 & 75.72 & 76.00 & 60.87 \\
\third{} \regt{} & 68.09 & & 74.79 & 76.89 & 77.26 & 61.28 & 68.82 & 69.10 & 54.55 \\

\eva{} & 90.66 & & 96.34 & 96.38 & 97.27 & 60.82 & 96.17 & 96.71 & 68.12 \\ 
\eva{} \regt{} & 90.66 & & 91.18 & 94.01 & 94.62 & 68.52 & 82.14 & 82.52 & 62.12 \\ 

\toprule
& \multicolumn{1}{c}{ImGT} & & \multicolumn{7}{c}{ImageNetMultiLabel~\cite{tsipras2020imagenetimageclassificationcontextualizing}} \\ \cmidrule{2-2} \cmidrule{4-10} 
& \multicolumn{1}{c}{$\mathrm{A_{10000}}$} & & \multicolumn{1}{c}{$\mathrm{{A{\setminus}N}_{9684}}$} & \multicolumn{1}{c}{$\mathrm{S_{7528}}$} & \multicolumn{1}{c}{$\mathrm{{S+}_{6201}}$} & \multicolumn{1}{c}{$\mathrm{{S-}_{1327}}$} & \multicolumn{1}{c}{$\mathrm{M_{2156}}$} & \multicolumn{1}{c}{$\mathrm{{M+}_{1999}}$} & \multicolumn{1}{c}{$\mathrm{{M-}_{157}}$} \\
\cmidrule{1-10}

\cgpt{} & 71.30 & & 73.85 & 72.74 & 82.04 & 29.31 & 77.74 & 80.74 & 39.49 \\
\cgpt{} \regt{} & 71.30 & & 77.36 & 79.05 & 83.25 & 58.39 & 73.29 & 74.52 & 57.52 \\

\qwen{} & 69.65 & & 73.72 & 71.24 & 80.63 & 27.35 & 82.37 & 84.64 & 53.50 \\
\qwen{} \regt{} & 69.65 & & 76.89 & 77.67 & 82.47 & 54.05 & 76.41 & 77.73 & 59.48 \\

\third{} & 68.31 & & 71.81 & 70.31 & 78.97 & 29.84 & 77.04 & 80.19 & 36.94 \\
\third{} \regt{} & 68.31 & & 74.52 & 75.88 & 80.20 & 54.63 & 72.15 & 73.60 & 53.59 \\

\eva{} & 90.82 & & 84.70 & 82.85 & 97.06 & 16.43 & 91.14 & 96.20 & 26.75 \\ 
\eva{} \regt{} & 90.82 & & 91.01 & 92.49 & 95.69 & 76.74 & 87.42 & 88.74 & 70.59 \\ 

\toprule
& \multicolumn{1}{c}{ImGT} & & \multicolumn{7}{c}{Label Errors~\cite{northcutt2021pervasivelabelerrorstest}} \\ \cmidrule{2-2} \cmidrule{4-10} 
& \multicolumn{1}{c}{$\mathrm{A_{5440}}$} & & \multicolumn{1}{c}{$\mathrm{{A{\setminus}N}_{4549}}$} & \multicolumn{1}{c}{$\mathrm{S_{3952}}$} & \multicolumn{1}{c}{$\mathrm{{S+}_{2554}}$} & \multicolumn{1}{c}{$\mathrm{{S-}_{1398}}$} & \multicolumn{1}{c}{$\mathrm{M_{597}}$} & \multicolumn{1}{c}{$\mathrm{{M+}_{597}}$} & \multicolumn{1}{c}{$\mathrm{{M-}_{0}}$} \\
\cmidrule{1-10}

\cgpt{} & 40.20 & & 59.90 & 57.69 & 57.91 & 57.30 & 74.54 & 74.54 & 0.00 \\
\cgpt{} \regt{} & 40.20 & & 63.64 & 64.79 & 64.98 & 64.45 & 72.59 & 72.59 & 0.00 \\

\qwen{} & 39.12 & & 58.65 & 55.74 & 55.68 & 55.87 & 77.89 & 77.89 & 0.00 \\
\qwen{} \regt{} & 39.12 & & 63.80 & 64.36 & 64.73 & 63.69 & 75.80 & 75.80 & 0.00 \\

\third{} & 38.90 & & 58.69 & 56.78 & 56.26 & 57.73 & 71.36 & 71.36 & 0.00 \\
\third{} \regt{} & 38.90 & & 61.89 & 63.42 & 63.49 & 63.30 & 69.35 & 69.35 & 0.00 \\

\eva{} & 63.24 & & 76.85 & 74.65 & 84.69 & 56.29 & 91.46 & 91.46 & 0.00 \\ 
\eva{} \regt{} & 63.24 & & 78.65 & 79.83 & 85.27 & 69.80 & 85.96 & 85.96 & 0.00 \\ 

\bottomrule
\end{tabular}
\label{tab:prior_work_ext}
\end{table*}

\begin{figure}[htbp]
    \centering
    \begin{tabular}{@{}cc@{}}
        \includegraphics[width=0.5\linewidth]{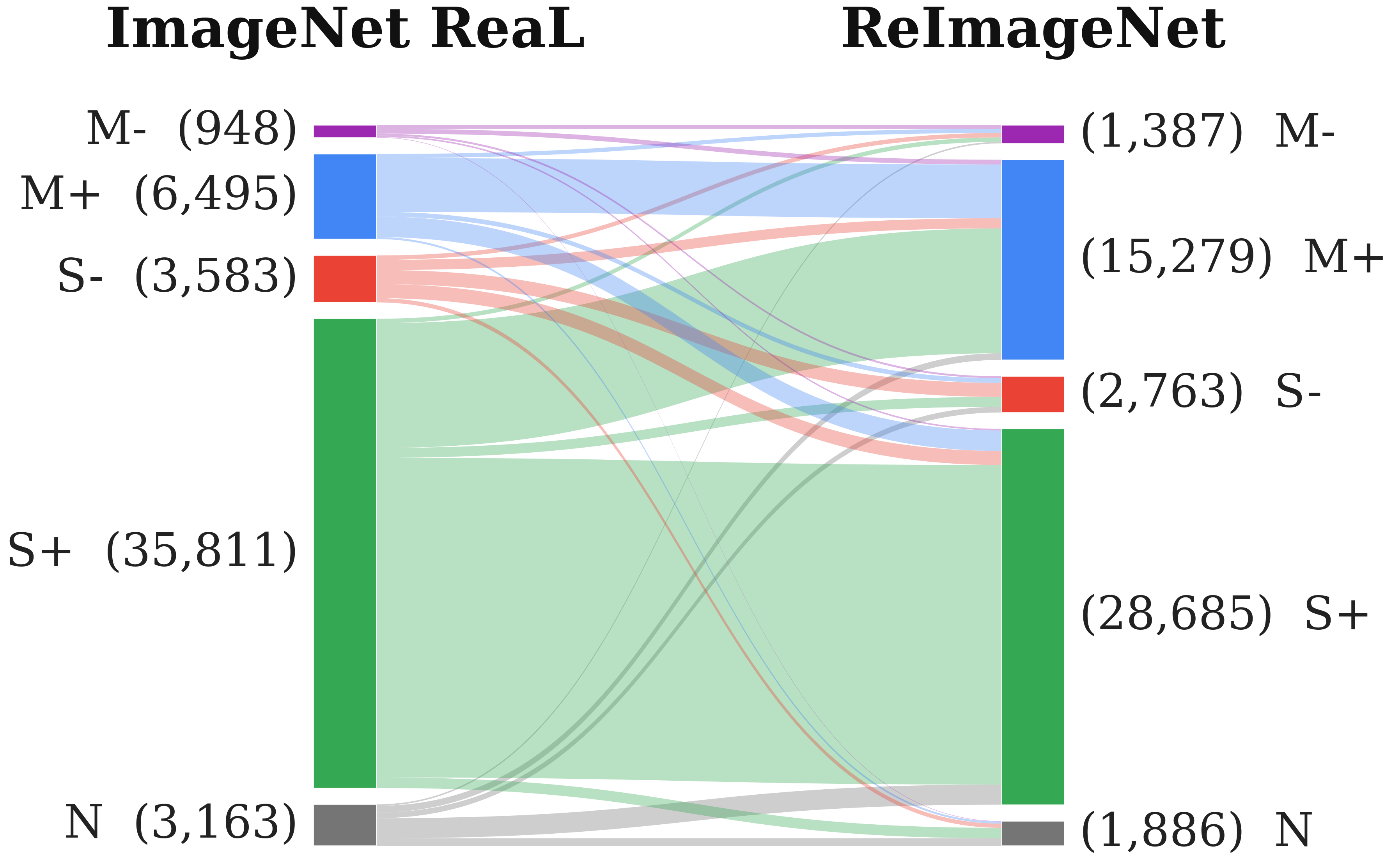} &
        \includegraphics[width=0.5\linewidth]{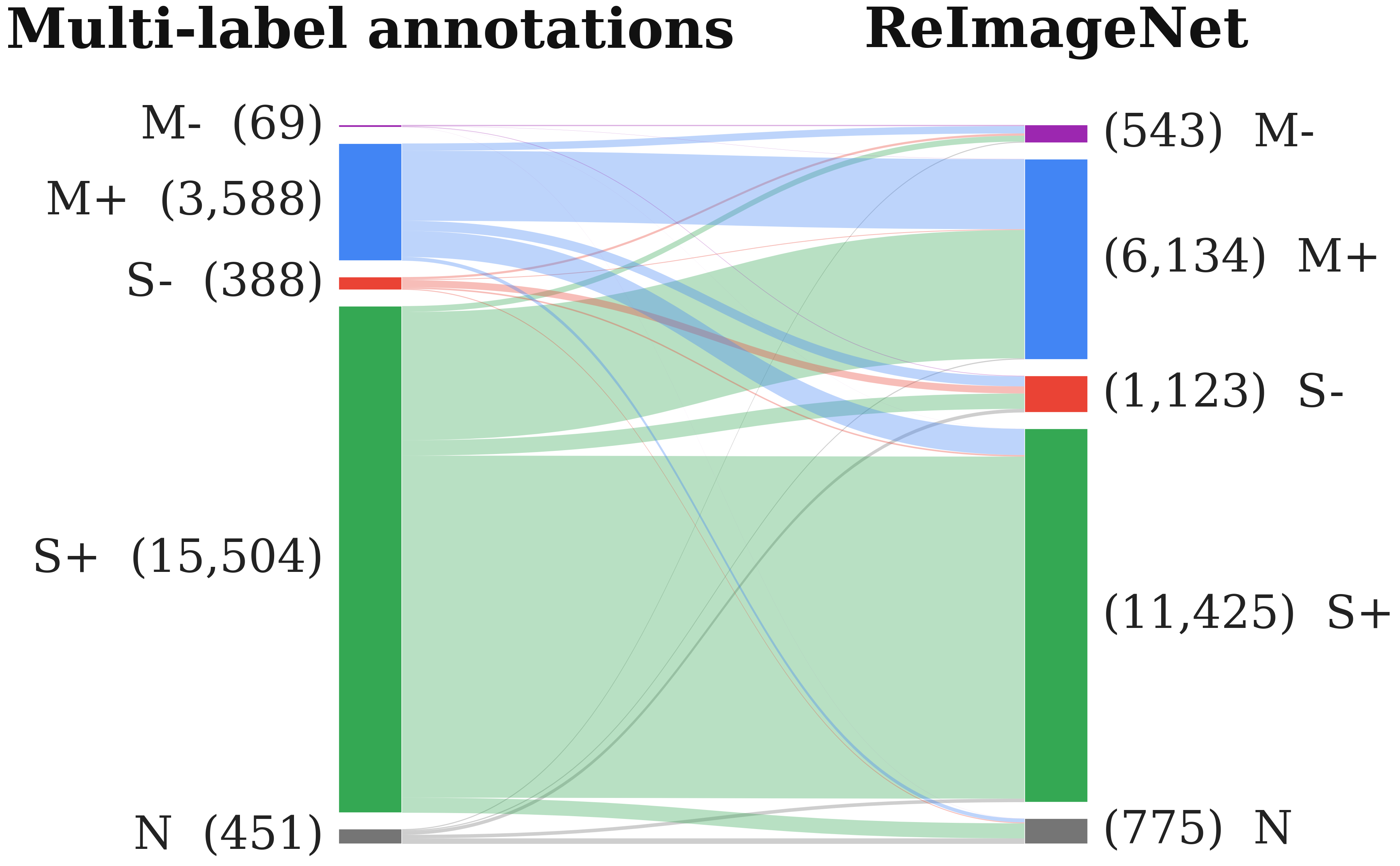} \\[20pt]
        \includegraphics[width=0.5\linewidth]{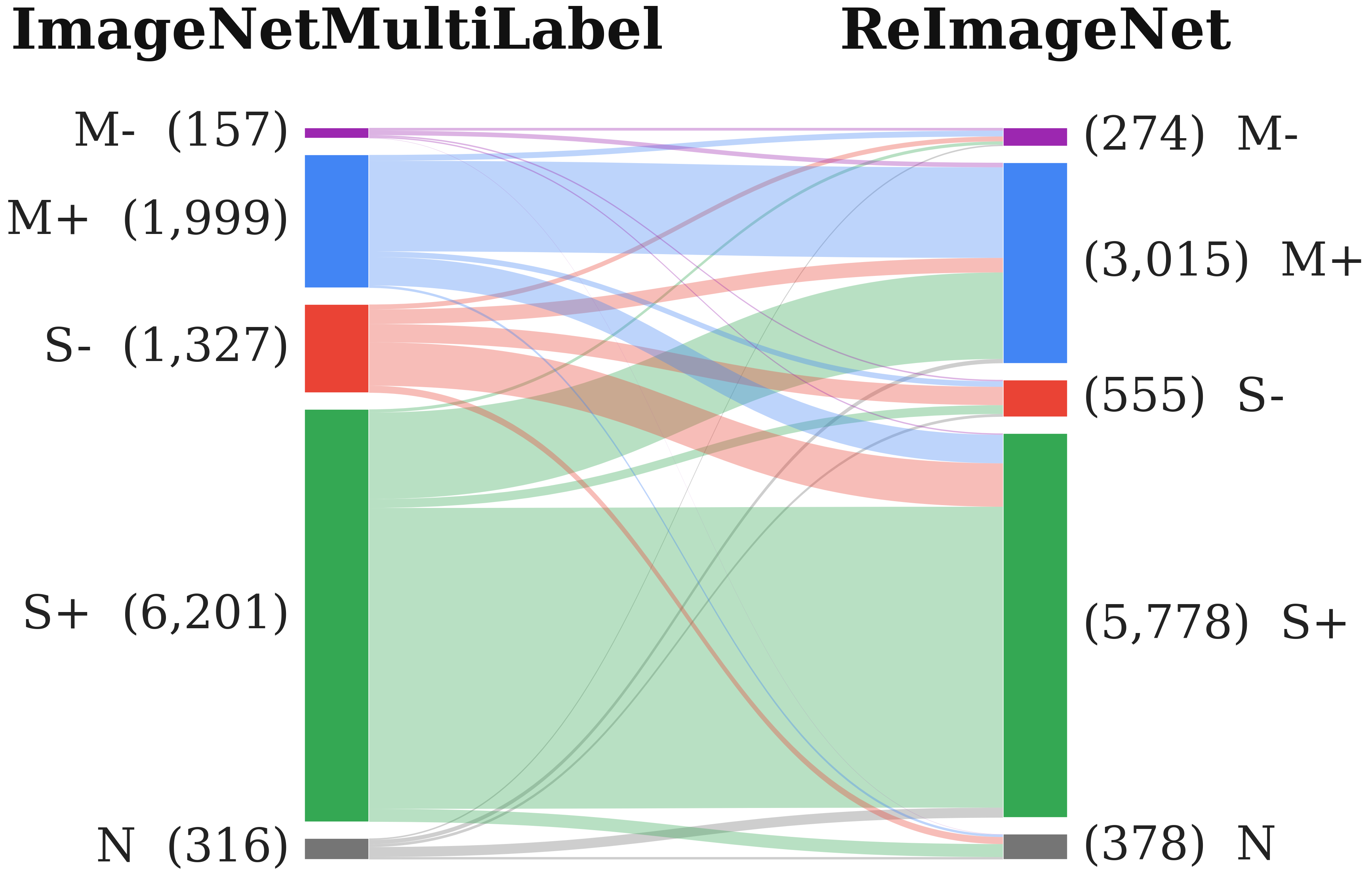} &
        \includegraphics[width=0.5\linewidth]{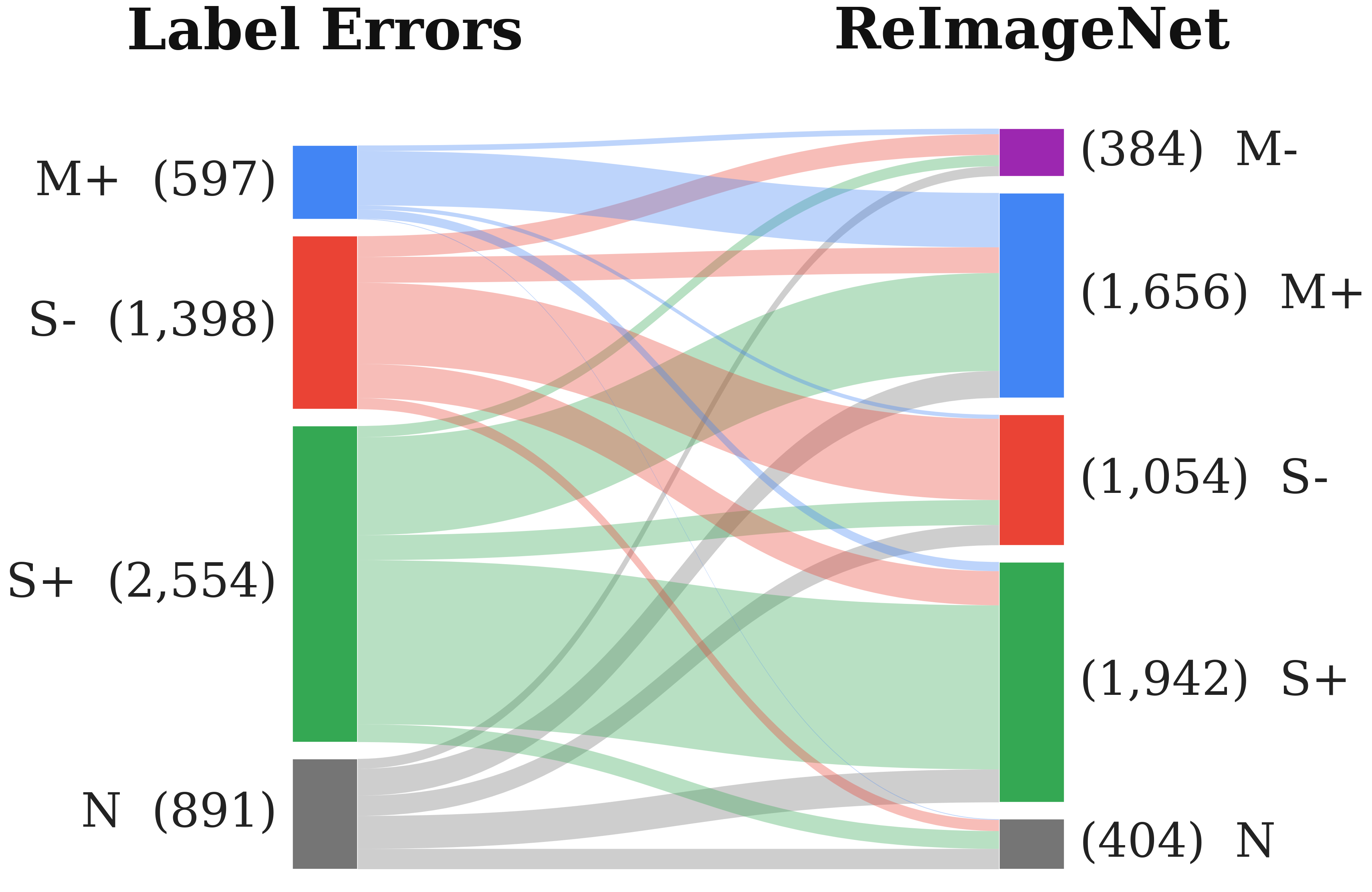} \\
    \end{tabular}
    \caption{Alluvial diagrams illustrating how proposed labels from prior \imnet{} reannotation efforts map to our \regt{} labels. Each image is assigned to a category as defined in~\Cref{fig:reannot_defs}, and the flows reveal the degree of agreement and disagreement.}
    \label{fig:ann_grid}
\end{figure}

We evaluate models against four existing reannotation efforts and compare the effect of applying our \regt{} label categories to each. 
Applying \regt{} raises accuracy on the challenging $\mathrm{S-}$ subsets on most datasets, by $20$–$60$\% on ReaL~\cite{beyer2020imagenet} and ImageNetMultiLabel~\cite{tsipras2020imagenetimageclassificationcontextualizing}, indicating that prior datasets left a substantial number of label errors or missing labels unaddressed. 
At the same time, $\mathrm{S+}$ and $\mathrm{M+}$ accuracy sometimes decreases under \regt{}, possibly
reflecting cases where our stricter definitions remove labels that 
prior works accepted. 
Direct comparison across datasets is not straightforward: the label distributions differ substantially (\eg $\mathrm{S-}$ ranges from $388$ to $3{,}583$ images, and Label 
Errors~\cite{northcutt2021pervasivelabelerrorstest} contains no 
$\mathrm{M-}$ images at all). A label-flow analysis is provided in~\Cref{fig:ann_grid}.

\clearpage
\newpage

\end{document}